\documentclass[11pt]{article}

\usepackage[final]{acl}

\usepackage{float}

\usepackage{times}
\usepackage{latexsym}
\usepackage{booktabs}
\usepackage{multirow}
\usepackage{amsfonts}
\usepackage{amsmath}
\usepackage{graphicx}
\usepackage{enumitem}
\usepackage{booktabs}
\usepackage{siunitx}
\usepackage{etoc}
\usepackage{hyperref}

\usepackage[T1]{fontenc}

\usepackage[utf8]{inputenc}

\usepackage{microtype}

\usepackage{inconsolata}

\usepackage{graphicx}

\title{Towards Efficient Post-Training via Fourier-Driven Adapter Architectures}

\author{
\textbf{Donggyun Bae}$^{1}$ 
\textbf{Jongil Park}$^{1,\dagger}$ \\
\vspace{0.5em}
$^{1}$Konkuk University \\
\vspace{0.5em}
$^{\dagger}$Corresponding author: \texttt{jipark@kkucc.konkuk.ac.kr}
}

\begin{document}
\maketitle
\begin{abstract}
We propose a novel framework, named Fourier-Activated Adapter (FAA) for parameter-efficient fine-tuning of larg-secale pre-trained language models. By integrating random Fourier features into the adapter module, our FAA decomposes input representations into high- and low-frequency components and employs a dynamic, frequency-aware activation mechanism to selectively emphasize crucial semantic signals. Our FAA improves the performance of the fine-tuned model and enhances the model's perception of multi-frequency semantic information. Experiments on GLUE, E2E NLG, and instruction tuning benchmarks demonstrate competitive or superior results of our FAA. Besides, ablation studies confirm the importance of frequency-aware activation and adaptive weighting. This demonstrates that our FAA is an effective and robust solution for enhancing the performance of large language models while maintaining computational efficiency.
\end{abstract}

\begin{figure*}[t]  
    \vspace{-1.5cm}
    \centering  
    \includegraphics[width=1\textwidth]{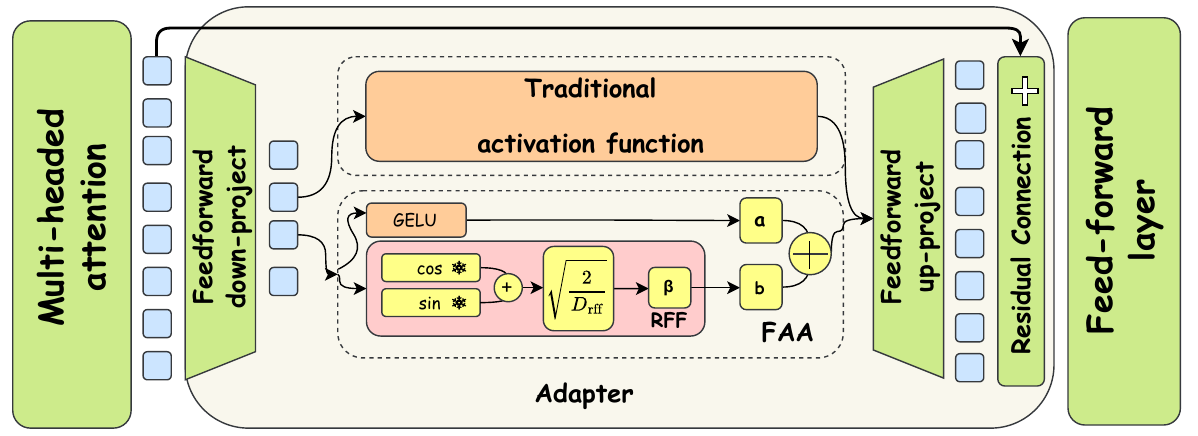}  
    \caption{Comparison of traditional adapters and Fourier-activated adapters (FAA). The top shows the traditional adapter, which consists of a feed-forward down-projection, a nonlinearity (e.g., a GELU function), and a feed-forward up-projection. The bottom shows the Fourier-activated adapter (FAA). The fixed activation function is replaced by RFF+GELU, where RFF first controls the consistency of the generated frequency features in the numerical scale through the normalization factor $\sqrt{\frac{2}{D_{\text{rff}}}}$, then controls the frequency distribution through the $\beta$ coefficient, and then combines with GELU through learnable coefficients $a$ and $b$ to achieve a frequency-aware Fourier activation strategy. Note that the adapter layers are interspersed between the attention layer and the feed-forward network.}
    \label{fig:example}  
\end{figure*}

\section{Introduction}
Large Language Models (LLMs) have become a fundamental technology in NLP, demonstrating exceptional language comprehension and generation capabilities\cite{brown2020languagemodelsfewshotlearners,z1}. With the rise of large-scale pre-trained models, LLMs have achieved remarkable progress in machine translation\cite{zhu2024multilingualmachinetranslationlarge,z2}, question answering\cite{bisk2019piqareasoningphysicalcommonsense,Z3,z4}, and text generation\cite{li2022pretrainedlanguagemodelstext,Z5}. However, they still face challenges in cross-domain generalization, out-of-distribution robustness\cite{yuan2023revisitingoutofdistributionrobustnessnlp,z6,z7}, and low-resource scenarios. When dealing with complex tasks or scarce data, fine-tuning large models directly not only requires extensive computational resources but also leads to suboptimal generalization, making it difficult to handle domain-specific terminology, intricate syntax, and sudden semantic shifts. To address these issues, Parameter-Efficient Fine-Tuning (PEFT) techniques\cite{dodge2020finetuningpretrainedlanguagemodels,xu2023parameterefficientfinetuningmethodspretrained} have been introduced. By freezing most of the base model’s parameters and only introducing lightweight adapter modules\cite{houlsby2019parameterefficienttransferlearningnlp} for fine-tuning, PEFT significantly reduces computational and storage costs while improving deployment efficiency. As an effective solution, PEFT enables LLMs to maintain high performance across various downstream tasks while minimizing computational demands. However, existing PEFT methods still struggle to capture high-frequency semantic information and handle complex tasks, limiting their effectiveness in cross-domain generalization and low-resource scenarios.

Current PEFT approaches face two major challenges. First, traditional adapter architectures are mostly designed with fixed activation functions\cite{hendrycks2023gaussianerrorlinearunits,z8}, making them rigid in responding to frequency variations in input data. This prevents them from dynamically adapting to high-frequency features in complex tasks, thereby affecting performance in specialized domains, intricate syntactic structures, and scenarios with rapid semantic changes. Second, while low-rank optimization methods (e.g., LoRA\cite{hu2021loralowrankadaptationlarge,z9,z10,z11}) achieve compression by reducing parameter count, they fail to fully leverage frequency-domain structures. As a result, these models struggle to capture fine-grained, high-frequency semantic features, making it difficult to maintain representation precision. Through an in-depth analysis of LLM behavior, we observe clear spectral sparsity in text processing\cite{tran2023sparsetopicmodelingspectral,z12,z13}, where core semantic information is often concentrated in a few key frequency bands, while most low-frequency components contribute little to meaningful representation. Previous works (\cite{verma2024signalprocessinglargelanguage,z14,z15}, \cite{leethorp2022fnetmixingtokensfourier,z16,z17}) have attempted to incorporate frequency-based enhancements into language models, yielding promising results. Meanwhile, the introduction of Kolmogorov-Arnold Networks (KAN)\cite{liu2025kankolmogorovarnoldnetworks} has provided new insights into learning adaptive activation functions. The subsequent development of Kolmogorov-Arnold Fourier Networks (KAF)\cite{zhang2025kolmogorovarnoldfouriernetworks,z18,z19} builds on KAN by integrating Fourier-based enhancements to improve spectral analysis capabilities.

Motivated by these findings, we propose a new approach that utilizes Random Fourier Features (RFF)\cite{NIPS2007_013a006f,z20,z21} to transform input signals into the frequency domain, reconfiguring the traditional adapter processing pipeline. Our goal is to design a frequency-domain processing module that introduces only a minimal parameter overhead to the base model while significantly enhancing generalization in cross-domain and low-resource scenarios. The proposed \textbf{Fourier-Activated Adapter Framework} consists of three key components. First, inspired by KA activation strategies, we construct a dynamic frequency-aware activation mechanism that allows the adapter to adjust its response to critical information dynamically across different tasks. Second, we introduce a randomized Fourier transform to decompose input signals in the frequency domain, leveraging spectral sparsity to effectively map high-frequency features into a lower-dimensional space. Finally, we develop a lightweight training strategy that incorporates sparsity constraints, reducing the parameter overhead introduced by Fourier transform while keeping the base model intact.

Experimental results demonstrate that our FAA Framework achieves significant improvements in cross-domain generalization and low-resource tasks. The proposed framework not only effectively reduces the number of parameters required for fine-tuning but also achieves performance comparable to or even surpassing traditional full fine-tuning methods across multiple tasks. By leveraging this novel technique, we provide an efficient and practical solution for deploying LLMs in real-world applications, particularly in resource-constrained environments and multi-task processing scenarios.
\section{Related Work}
\subsection{Parameter-Efficient Fine-Tuning Techniques}
Parameter-Efficient Fine-Tuning (PEFT) methods have gained widespread application in the adaptive adjustment of large-scale pre-trained language models in recent years. Traditional full-parameter fine-tuning methods\cite{liu2019multitaskdeepneuralnetworks,lv2024parameterfinetuninglargelanguage,han2016deepcompressioncompressingdeep,z22} require updating a large number of model parameters when dealing with specific tasks, leading to high computational and storage costs. To address this issue, researchers have proposed various PEFT methods, such as Adapters \cite{houlsby2019parameterefficienttransferlearningnlp} and LoRA \cite{hu2021loralowrankadaptationlarge}. Adapters insert lightweight adapter modules between the layers of the model, fine-tuning only these new parameters, thereby significantly reducing the number of parameters required for fine-tuning. LoRA further reduces the scale of parameter updates through low-rank matrix decomposition. These methods improve the efficiency and flexibility of fine-tuning while maintaining model performance.
\subsection{Frequency Domain Enhancement and Fourier Transform}
Frequency domain analysis, successful in computer vision\cite{192463,xu2020learningfrequencydomain}, is gaining traction in NLP\cite{verma2024signalprocessinglargelanguage}. By transforming text signals into the frequency domain, these methods better capture high and low-frequency features, improving pattern understanding. Recent work\cite{He_2023,hua2025fourierpositionembeddingenhancing} has integrated Fourier transforms into language models, enhancing multi-frequency semantic modeling\cite{jin2024languagemodelssemanticindexers} and showing benefits in cross-domain and low-resource scenarios.

Studies\cite{+2012,tamkin2020languageprismspectralapproach} show that key semantic information concentrates in specific frequency bands, with methods like FourierFT\cite{gao2024parameterefficientfinetuningdiscretefourier} decomposing inputs to better capture multi-frequency components. However, current approaches have not fully leveraged frequency domain structures for semantic representation, making the optimization of these techniques an important research direction.
\subsection{Kolmogorov-Arnold Networks (KAN) and Fourier Activation}
Kolmogorov-Arnold Networks (KAN)\cite{liu2025kankolmogorovarnoldnetworks} introduced a new activation mechanism that improves model response through adaptive learning. This evolution led to Kolmogorov-Arnold Fourier Networks (KAF)\cite{zhang2025kolmogorovarnoldfouriernetworks}, which combines Fourier transforms with a frequency-aware activation mechanism, allowing dynamic adjustment to different frequency information and better capture of high-frequency details.

Building on these advances, we propose a Fourier-Activated Adapter framework (FAA) based on random Fourier features, aiming to enhance large language models' performance in cross-domain generalization and low-resource scenarios while maintaining efficient parameter updates.
\section{Methodology}
Traditional adapter modules are lightweight components inserted into pre-trained models for task adaptation via Parameter-Efficient Fine-Tuning (PEFT). They compress input features into a lower-dimensional space and then reconstruct them back to the original dimension using a down-projection and up-projection. A nonlinear activation function (e.g., ReLU) is applied between the projections for enhanced expressiveness.
\begin{equation}
h_{\text{adapter}}^{(l)} = h^{(l)} + W_{\text{up}}^{(l)} \cdot \sigma(W_{\text{down}}^{(l)} \cdot h^{(l)} + b_{\text{down}}^{(l)}) + b_{\text{up}}^{(l)}
\end{equation}
Where:\( W_{\text{down}}^{(l)} \in \mathbb{R}^{r \times d_{\text{model}}} \) is the down-projection matrix.\( W_{\text{up}}^{(l)} \in \mathbb{R}^{d_{\text{model}} \times r} \) is the up-projection matrix.
\( b_{\text{down}}^{(l)} \) and \( b_{\text{up}}^{(l)} \) are learnable bias terms.\( \sigma(\cdot) \) is a nonlinear activation function.
We freeze the original model parameters and only update the adapter parameters \( \theta_{\text{adapter}} = \{ W_{\text{down}}, W_{\text{up}}, b_{\text{down}}, b_{\text{up}} \} \).
\subsection{Fourier Activation Adapter and Frequency Response Enhancement}
Traditional activation functions struggle with modeling frequency-domain features. To address this, we propose the Fourier Activation Adapter (FAA), which integrates Random Fourier Features (RFF) into the adapter module, enhancing the model's ability to capture multi-frequency semantic components. This is particularly useful for fine-tuning, as it allows the model to better capture complex patterns at multiple scales.
\subsubsection{Random Fourier Feature Transformation and Frequency-Aware Activation Mechanism}
In the FAA framework, a dual-channel Random Fourier Feature (RFF) transformation is applied to the input \( h^{(l)} \in \mathbb{R}^{d_{\text{model}}} \) of the \( l \)-th layer. The transformation is given by:
\begin{equation}
\begin{aligned}
z_{\text{RFF}}^{(l)} = \sqrt{\frac{2}{D_{\text{rff}}}} \left[ \cos\left( W_{\text{rff}}^{(l) \top} h^{(l)} + b_{\text{rff}}^{(l)} \right) \right. \\
\left. \oplus \sin\left( W_{\text{rff}}^{(l) \top} h^{(l)} + b_{\text{rff}}^{(l)} \right) \right]
\end{aligned}
\label{RFF}
\end{equation} 
Where \( W_{\text{rff}}^{(l)} \in \mathbb{R}^{d_{\text{model}} \times D_{\text{rff}}} \) is drawn from a Gaussian distribution \( N(0, \sigma^{-2}) \), \( b_{\text{rff}}^{(l)} \in \mathbb{R}^{D_{\text{rff}}} \) is drawn from \( U(0, 2\pi) \), and \( \oplus \) denotes concatenation of cosine and sine terms. The parameter \( \sigma \) controls the bandwidth, where smaller \( \sigma \) captures high-frequency signals (e.g., edges), and larger \( \sigma \) captures low-frequency signals (e.g., global structures).
The feature transformation is then fused with a frequency-aware activation mechanism:
\begin{equation}
h^{(l)} = \alpha^{(l)} \odot \text{GELU}\left( W_{\text{base}}^{(l)} h^{(l)} \right) + \beta^{(l)} \odot z_{\text{RFF}}^{(l)}
\label{eq:ab}
\end{equation}
Where \( W_{\text{base}}^{(l)} \in \mathbb{R}^{d_{\text{model}} \times d_{\text{model}}} \) is a learnable projection matrix, \( \text{GELU}(\cdot) \) is the non-linear activation function, and \( \alpha^{(l)}, \beta^{(l)} \in \mathbb{R}^{d_{\text{model}}} \) are learnable channel attention vectors. The dynamic weights adjust the fusion of time-domain and frequency-domain features, with \( \odot \) denoting the element-wise Hadamard product.
This method ensures both time-domain and frequency-domain information are captured for more flexible and robust feature representation. The random projection matrix \( W_{\text{rff}} \) can be frozen during optimization, making the method suitable for low-resource settings.
\subsubsection{Dynamic Enhancement of High-Frequency Features}
In deep neural networks, high-frequency features often decay, leading to ineffective propagation of local information (e.g., syntactic boundaries, texture features). To address this, we propose a Frequency-Responsive Gating Mechanism that uses adaptive spectral analysis to selectively enhance high-frequency components and suppress interference from irrelevant frequency bands.
We first decompose the input signal $h^{(l)}$ into $n$ frequency channels by projecting it onto cosine and sine components:
\vspace{-5pt} 
\begin{equation}
\scalebox{0.8}{$
\begin{aligned}
g_i^{(l)} &= \cos\left( w_{i}^{(l) \top} h^{(l)} + b_{i}^{(l)} \right) \oplus \sin\left( w_{i}^{(l) \top} h^{(l)} + b_{i}^{(l)} \right) \\
&\in \mathbb{R}^{2d_{\text{model}}} \quad (i=1,...,n)
\end{aligned}
$}
\end{equation}
\vspace{-5pt} 
Where: $w_i^{(l)} \in \mathbb{R}^{d_{\text{model}}}$ is the projection vector for the $i$-th frequency component, and $b_i^{(l)}$ is the phase offset. The concatenation of cosine and sine functions ensures each frequency channel captures both amplitude and phase information.
Next, we assign adaptive gating weights $r_i^{(l)}$ to each frequency component, with a Frequency-Responsive Gating mechanism that computes the weighted aggregation of frequency channels:
\begin{equation}
z_{\text{RFF}}^{(l)} = \sum_{i=1}^{n} r_i^{(l)} \cdot \text{LayerNorm}(g_i^{(l)})
\end{equation}
Where: $r_i^{(l)}$ controls the contribution of each frequency channel, and $\text{LayerNorm}(\cdot)$ normalizes the output to prevent instability. This mechanism allows the model to dynamically adjust its focus on different frequency bands, depending on task-specific requirements.
\subsection{Adaptive Frequency Weight Adjustment and Training Objective}
In deep learning models, different tasks require varying emphasis on high-frequency and low-frequency information, so fixed frequency weights may not adapt well to diverse data distributions. To address this, we propose an Adaptive Frequency Weight Adjustment Mechanism, which uses hierarchical gating weights to dynamically adjust frequency components and incorporates sparsity regularization to enhance frequency selection accuracy.
\subsubsection{Adaptive Frequency Weight Adjustment}
To improve the model's ability to perceive different frequency components, we design a hierarchical gating mechanism that allows dynamic weight adjustment for each frequency component at different layers. The weight $r_i^{(l)}$ for the $i$-th frequency component at layer $l$ is computed as:
\begin{equation}
\scalebox{0.8}{$
\begin{aligned}
r_i^{(l)} &= \sigma\left(a_i^{(l)} \odot \tilde{z}_i^{(l)} + c_i^{(l)}\right) \\
\tilde{z}_i^{(l)} &= \frac{1}{d_{\text{model}}} \sum_{k=1}^{d_{\text{model}}} W_{\text{gate},i}^{(l)} h_k^{(l)}
\end{aligned}
$}
\label{eq:Adaptive Frequency Weight Adjustment}
\end{equation}
Where: $a_i^{(l)} \in \mathbb{R}^{d_{\text{model}}}$: frequency sensitivity coefficients, controlling the influence of input features on each frequency; $c_i^{(l)} \in \mathbb{R}$: gating bias term, adding flexibility to adapt to different frequency distributions at each layer; $W_{\text{gate},i}^{(l)} \in \mathbb{R}^{d_{\text{model}} \times d_{\text{model}}}$: feature projection matrix, projecting input features onto the frequency channel; $\sigma(\cdot)$: Sigmoid activation function, ensuring that $r_i^{(l)}$ is bounded within $[0, 1]$, interpretable as the probability of selecting the frequency.
\subsubsection{Sparsity Constraints and Training Objective}
To improve frequency selectivity and avoid redundant computations, we introduce a dual regularization objective: L1 sparsity regularization to encourage sparse selection of important frequencies, and orthogonality penalties to reduce redundant interactions between frequency channels. The loss function is:
\begin{equation}
\scalebox{0.8}{$
\begin{aligned}
\mathcal{L}_{\text{freq}} &= \sum_{l=1}^{L} \left( \lambda_1 \sum_{i=1}^{n} |r_i^{(l)}|_1 + \lambda_2 \sum_{i<j} |r_i^{(l)} \circ r_j^{(l)}|_2^2 \right) \\
\mathcal{L}_{\text{total}} &= \mathcal{L}_{\text{task}} + \mathcal{L}_{\text{freq}}
\end{aligned}
$}
\label{L1}
\end{equation}
Where: $\mathcal{L}{\text{task}}$: is the base task loss (e.g., cross-entropy in NLP tasks); $\mathcal{L}{\text{freq}}$: is the frequency regularization loss, controlling the accuracy of frequency selection.
Through the Adaptive Frequency Weight Adjustment and Sparsity Regularization, this method improves the model's frequency domain modeling in multiple ways. First, hierarchical gating weights allow independent adjustment of high- and low-frequency information, improving task adaptability. Second, L1 regularization retains only the most important frequency components, reducing computational redundancy and improving generalization, while orthogonality penalties prevent redundant interactions between frequency channels, ensuring complementary frequency spectrum information and optimizing feature representation.
\subsection{FAA Integration with Base Models}
FAA adopts a modular design, allowing seamless integration with pre-trained language models (e.g., RoBERTa, BERT, LLM) in the Transformer architecture. This section formalizes the integration strategy to ensure compatibility with base models while maintaining efficient parameter utilization and supporting flexible fine-tuning.

\subsubsection{Structural Integration and Information Flow Modeling}
After integrating the FAA module, the information flow of the $l$-th layer Transformer is adjusted as follows:
\begin{equation}
\begin{aligned}
h_{\text{attn}}^{(l)} &= \text{MultiHeadAttn}(h^{(l-1)}) \\
h_{\text{FAA}}^{(l)} &= \text{FAA}(h_{\text{attn}}^{(l)}) \quad (\text{adapter module}) \\
h_{\text{mid}}^{(l)} &= \text{LayerNorm}(h^{(l-1)} + h_{\text{attn}}^{(l)} + \gamma^{(l)} \odot h_{\text{FAA}}^{(l)}) \\
h_{\text{ffn}}^{(l)} &= \text{FFN}(h_{\text{mid}}^{(l)}) \\
h^{(l)} &= \text{LayerNorm}(h_{\text{mid}}^{(l)} + h_{\text{ffn}}^{(l)})
\end{aligned}
\end{equation}
FAA is inserted in parallel after the self-attention output $h_{\text{attn}}^{(l)}$, without altering the base Transformer flow. The gating coefficient $\gamma^{(l)} \in \mathbb{R}^{d_{\text{model}}}$ is a learnable vector controlling FAA’s contribution, activating only when needed. The residual connection $h^{(l-1)} + h_{\text{attn}}^{(l)}$ ensures stable gradient flow and training stability.
\begin{table*}[ht]
    \vspace{-1.5cm}
    \centering
    \small
    \begin{tabular}{l l c c c c c c c c}
    \toprule
    \multirow{2}{*}{Method} & \multirow{2}{*}{\#Paras} & \multicolumn{8}{c}{Datasets} \\
    \cmidrule{3-10}
    & & CoLA & SST-2 & MRPC & QQP & QNLI & RTE & STS-B & WNLI \rule{0pt}{8pt} \\
    & & (MCC) & (Acc.) & (Acc.) & (Acc.) & (Acc.) & (Acc.) & (PCC) & (PCC) \\
    \midrule
    FF & 125M & 63.6\textsubscript{±0.4} & 94.8\textsubscript{±0.2} & 90.2\textsubscript{±0.1} & 93.2\textsubscript{±0.4} & 92.8\textsubscript{±0.5} & \textbf{81.5}\textsubscript{±0.2} & 91.2\textsubscript{±0.8} & 65.1\textsubscript{±0.1} \\
    AdapterH & 0.6M & 60.8\textsubscript{±0.4} & 94.2\textsubscript{±0.1} & 88.5\textsubscript{±1.1} & 93.5\textsubscript{±0.3} & 93.1\textsubscript{±0.1} & 71.5\textsubscript{±1.2} & 89.7\textsubscript{±0.3} & 64.2\textsubscript{±0.5} \\
    AdapterL & 0.6M & 62.6\textsubscript{±0.9} & 94.7\textsubscript{±0.3} & 88.4\textsubscript{±0.1} & \textbf{94.8}\textsubscript{±0.2} & 93.0\textsubscript{±0.2} & 75.9\textsubscript{±0.2} & 90.3\textsubscript{±0.1} & 64.5\textsubscript{±0.3}  \\
    AdapterP & 0.3M & 63.4\textsubscript{±1.2} & 95.1\textsubscript{±0.2} & 89.7\textsubscript{±0.7} & 93.0\textsubscript{±0.5} & 93.3\textsubscript{±0.3} & 78.4\textsubscript{±0.8} & 91.5\textsubscript{±0.2} & 65.0\textsubscript{±0.4}\\
    Compacter & 0.3M & 62.0\textsubscript{±0.6} & 94.5\textsubscript{±0.2} & 88.7\textsubscript{±0.5} & 92.3\textsubscript{±0.4} & 93.1\textsubscript{±0.2} & 81.0\textsubscript{±0.6} & 90.5\textsubscript{±0.2} & 64.8\textsubscript{±0.2}  \\
    Parallel Adapter & 1.2M & 61.1\textsubscript{±0.3} & 94.3\textsubscript{±0.5} & 89.5\textsubscript{±0.5} & 94.7\textsubscript{±0.4} & 92.2\textsubscript{±0.5} & 78.7\textsubscript{±0.7} & 91.1\textsubscript{±0.6} & 64.9\textsubscript{±0.1}  \\
    LoRA & 0.3M & \textbf{63.8}\textsubscript{±1.6} & 94.2\textsubscript{±0.3} & 90.0\textsubscript{±0.8} & 93.5\textsubscript{±0.6} & 92.2\textsubscript{±0.1} & 79.1\textsubscript{±0.5} & \textbf{92.8}\textsubscript{±0.4} & 65.2\textsubscript{±0.3}  \\
    FourierFT & 0.024M & 62.3\textsubscript{±1.4} & 94.2\textsubscript{±0.2} & 90.3\textsubscript{±0.3} & 92.0\textsubscript{±0.4} & 91.7\textsubscript{±0.4} & 78.4\textsubscript{±1.6} & 91.0\textsubscript{±0.4} & 66.0\textsubscript{±0.5} \\
    FAA (Ours) & 0.6M & 63.3\textsubscript{±0.4} & \textbf{96.1}\textsubscript{±0.1} & \textbf{91.0}\textsubscript{±0.7} & 94.5\textsubscript{±0.2} & \textbf{93.7}\textsubscript{±0.1} & 80.0\textsubscript{±0.2} & 91.4\textsubscript{±1.3} & \textbf{67.0}\textsubscript{±0.6} \\
    \midrule
    FF & 356M & 68.0\textsubscript{±0.1} & \textbf{96.4}\textsubscript{±0.5} & 90.9\textsubscript{±0.7} & 92.0\textsubscript{±0.6} & \textbf{95.0}\textsubscript{±0.3} & 86.6\textsubscript{±0.8} & \textbf{92.4}\textsubscript{±0.3} & 66.5\textsubscript{±0.5} \\
    AdapterH & 1.8M & 68.3\textsubscript{±1.0} & 96.1\textsubscript{±0.3} & 90.2\textsubscript{±0.7} & 91.8\textsubscript{±0.5} & 94.8\textsubscript{±0.2} & 83.8\textsubscript{±2.9} & 92.1\textsubscript{±0.7} & 65.5\textsubscript{±0.3} \\
    AdapterL & 1.8M & 67.8\textsubscript{±2.5} & 96.6\textsubscript{±0.2} & 89.7\textsubscript{±1.2} & 91.5\textsubscript{±0.4} & 94.8\textsubscript{±0.3} & 80.1\textsubscript{±2.9} & 91.9\textsubscript{±0.4} & 65.8\textsubscript{±0.2} \\
    AdapterP & 0.9M & 66.5\textsubscript{±0.4} & 96.2\textsubscript{±0.3} & 88.7\textsubscript{±2.9} & 91.2\textsubscript{±0.6} & 94.7\textsubscript{±0.2} & 83.4\textsubscript{±1.1} & 91.0\textsubscript{±1.7} & 65.3\textsubscript{±0.4}  \\
    Compacter & 0.9M & 66.3\textsubscript{±2.0} & 96.3\textsubscript{±0.5} & 87.7\textsubscript{±1.7} & 91.0\textsubscript{±0.5} & 94.7\textsubscript{±0.2} & \textbf{88.4}\textsubscript{±2.9} & 91.5\textsubscript{±0.5} & 65.0\textsubscript{±0.3}  \\
    Parallel Adapter & 4.8M & 68.2\textsubscript{±1.9} & 96.2\textsubscript{±0.5} & 90.2\textsubscript{±1.0} & 91.8\textsubscript{±0.4} & 94.8\textsubscript{±0.3} & 85.2\textsubscript{±1.1} & 92.3\textsubscript{±0.5} & 66.0\textsubscript{±0.2}  \\
    LoRA & 0.8M & 67.1\textsubscript{±1.4} & 96.0\textsubscript{±0.2} & \textbf{91.5}\textsubscript{±0.3} & 91.5\textsubscript{±0.4} & 94.4\textsubscript{±0.4} & 87.4\textsubscript{±1.6} & 91.9\textsubscript{±0.4} & 66.2\textsubscript{±0.3} \\
    FourierFT & 0.048M & 68.5\textsubscript{±1.2} & 95.3\textsubscript{±0.3} & 91.2\textsubscript{±0.4} & 92.0\textsubscript{±0.5} & 94.9\textsubscript{±0.3} & 87.5\textsubscript{±1.4} & 92.5\textsubscript{±0.5} & 66.8\textsubscript{±0.4}\\
    FAA (Ours) & 1.8M & \textbf{69.0}\textsubscript{±0.1} & 96.0\textsubscript{±0.9} & 90.0\textsubscript{±0.5} & \textbf{92.8}\textsubscript{±0.7} & 94.7\textsubscript{±0.8} & 86.2\textsubscript{±0.4} & 91.8\textsubscript{±0.2} & \textbf{67.5}\textsubscript{±0.2} \\
    \bottomrule
    \end{tabular}
    \caption{Performance of various fine-tuning methods with RoBERTa Base (upper part) and RoBERTa Large (lower part) models on 8 datasets of the GLUE benchmark. We report the Matthew's correlation coefficient (MCC) for CoLA, Pearson correlation coefficient (PCC) for STS-B and WNLI, and accuracy (Acc.) for all the remaining tasks. We report the median result of 5 runs, each using different random seeds. The best results for each dataset are shown in bold. Higher is better for all metrics in 8 datasets.}
    \label{tab:glue_results}
\end{table*}
\subsubsection{Parameter Freezing and Fine-Tuning Strategy}
To improve the parameter efficiency of FAA, we adopt a staged update strategy, updating only the FAA-related parameters while freezing the base Transformer parameters. Let the full model parameter set be:
\begin{equation}
\Theta = \underbrace{\{ \theta_{\text{attn}}^{(l)}, \theta_{\text{ffn}}^{(l)} \}_{l=1}^L}_{\Theta_{\text{base}}} \cup \underbrace{\{ \theta_{\text{FAA}}^{(l)}, \gamma^{(l)} \}_{l \in \mathcal{L}_{\text{insert}}} }_{\Theta_{\text{FAA}}}
\end{equation}
Where: \( \Theta_{\text{base}} \): Transformer backbone parameters, including the self-attention module and feed-forward network (FFN). \( \Theta_{\text{FAA}} \): FAA-related parameters, including frequency modeling weights \( \theta_{\text{FAA}}^{(l)} \) and gating vectors \( \gamma^{(l)} \).During fine-tuning, only the FAA-related parameters are optimized, while the base Transformer parameters are frozen:
Frozen parameters : $\frac{\partial \mathcal{L}}{\partial \Theta_{\text{base}}} = 0$.Updated parameters: $\Theta_{\text{FAA}} \leftarrow \Theta_{\text{FAA}} - \eta \nabla_{\Theta_{\text{FAA}}} \mathcal{L}$Modular extension characteristics, the mathematical form of FAA supports multi-dimensional extension: 
\textbf{Hierarchical Heterogenization}: Different layers can configure independent hyperparameters: \( \theta_{\text{FAA}}^{(l)} = \{ D_{\text{rff}}^{(l)}, \sigma^{(l)}, n^{(l)} \} \)
\textbf{Multi-modal Extension}: For vision language models, define cross-modal frequency projections: $ W_{\text{rff, cross}}^{(l)} = [W_{\text{text}}^{(l)} | W_{\text{image}}^{(l)}] \in \mathbb{R}^{(d_t+d_v) \times D_{\text{rff}}} $ \textbf{ dynamic topology}: Implement conditional computation via a gating mechanism: 
$
\gamma^{(l)} = \text{Sigmoid}(W_{\text{gate}}^{(l)} h_{\text{attn}}^{(l)})
$
This design ensures that the FAA module is backward compatible, as when \( \gamma^{(l)} \to 0 \), it degenerates into a standard Transformer. It also supports forward compatibility, allowing for the subsequent addition of spectrum normalization and other extensions. Additionally, the FAA module exhibits cross-architecture universality, as it imposes no special constraints on \( d_{\text{model}} \), making it adaptable to models of various sizes.

\section{Experiments}

We evaluate FAA fine-tuned NLP models across three perspectives: (1) NLU(natural language understanding) tasks on the GLUE benchmark\cite{wang2019gluemultitaskbenchmarkanalysis} with RoBERTa (Base \& Large)\cite{liu2019robertarobustlyoptimizedbert}, (2) NLG(natural language generation) tasks on the E2E NLG dataset\cite{Du_ek_2020} using GPT2-Small\cite{radford2019language}, DeepSeek-R1-Distill-Qwen-1.5B\cite{deepseekai2025deepseekr1incentivizingreasoningcapability}, LLaMA2-7B\cite{touvron2023llama2openfoundation}, and LLaMA3-8B\cite{grattafiori2024llama3herdmodels}, and (3) instruction tuning tasks on MT-Bench\cite{zheng2023judgingllmasajudgemtbenchchatbot}, Vicuna Eval\cite{chiang2023vicuna}, BBH\cite{suzgun2022challenging}, MATH\cite{hendrycksmath2021}, and Alpaca\cite{alpaca} with DeepSeek-R1-Distill-Qwen-1.5B, LLaMA2-7B, Qwen2-7B, and LLaMA3-8B.For a detailed introduction to the dataset, see the Supplementary Materials section in the Appendix \ref{APP:Datasets and Tasks Overview}.In addition, we also designed frequency perception experiments and ablation experiments to test the specific frequency performance of the FAA fine-tuned model and the impact of each component on the FAA model.All experiments were performed on A100 64G.

\subsection{Compared PEFT Methods}

We compare the FAA method with currently popular parameter-efficient fine-tuning (PEFT) methods, using the experimental settings of each respective method. The models involved in the comparison include:\textbf{Full Parameter Fine-tuning (FF)}: All parameters are updated, leading to high computational and storage costs. • \textbf{AdapterH}: Inserts an adapter layer between self-attention and the feedforward network. • \textbf{AdapterL\cite{adapterl}}: Adds a lightweight adapter layer only after the MLP module. • \textbf{AdapterP}: Optimizes adapter placement after the feedforward layer for better task adaptation. • \textbf{Compacter\cite{mahabadi2021compacterefficientlowrankhypercomplex}}: Uses low-rank parameterization to reduce storage and computation. • \textbf{Parallel Adapter\cite{huh2024trainingneuralnetworksscratch}}: Uses parallel adapters to enhance inference efficiency. • \textbf{LoRA}: Fine-tunes low-rank matrices to reduce the parameter updates during training. • \textbf{FourierFT}: Replaces low-rank approximations with Fourier transforms to cut down parameters.
Please note that due to model adaptation and dataset loading issues, we may choose different comparison models for different tasks.

\subsection{Natural Language Understanding}
\subsubsection{Experimental Setup}
The baseline models are pre-trained RoBERTa Base (12 layers, 768 hidden units) and RoBERTa Large (24 layers, 1024 hidden units), using their official configurations. During fine-tuning, we adopt FAA (Feature-wise Attention Adapter) as the adapter layer, which is inserted between the Transformer layers and feed-forward layers, with a total of 4 adapter layers. Additionally, the weights of all structures, except for the classification head, are frozen during fine-tuning. The specific hyperparameter settings for the experiments are provided in Appendix ~\ref{APP:Hyperparameter settings}. We evaluate the fine-tuned models on their comprehension ability across eight tasks: CoLA, SST-2, MRPC, QQP, QNLI, RTE, STS-B, and WNLI. For specific training time comparisons, see the supplementary materials section in the appendix~\ref{APP:Training Time Analysis}.
\subsubsection{Experimental Results}
Table~\ref{tab:glue_results} shows that FAA outperforms other methods on multiple GLUE tasks, such as CoLA (MCC: 63.3), WNLI (PCC: 67.0), and QQP (Acc.: 94.5). Compared with traditional adapters, FAA achieves strong performance, indicating that it has better performance for fine-tuning language models and has a more robust effect on NLU tasks.
\subsection{Natural Language Generation}
\subsubsection{Experimental Setup}
We evaluate the natural language generation capability of FAA fine-tuned models on the E2E NLG task, using GPT2-Small, DeepSeek-R1-Distill-Qwen-1.5B, LLaMA2-7B, and LLaMA3-8B. Models are evaluated using BLEU, NIST, METEOR, ROUGE-L, and CIDEr. The models are trained for 30 epochs, and results are recorded from the best test set performance. Specific hyperparameter settings are detailed in Appendix ~\ref{APP:Hyperparameter settings}.
\begin{table}
\small
\setlength{\tabcolsep}{2pt} 
\caption{Performance comparison of different methods on the End-to-End Natural Language Generation Benchmark using BLEU, NIST, METEOR, ROUGE-L, and CIDEr for GPT-2 Small, Deepseek R1-1.5B, LLaMA2-7B and LLaMA3-8B. We ran 10 experiments with different random seeds and recorded the best test set performance.}
\label{tab:e2e}
\resizebox{\columnwidth}{!}{%
\begin{tabular}{@{}c|l|r|ccccc@{}}
\toprule
Model & Method & \multicolumn{1}{c|}{\begin{tabular}[c]{@{}c@{}}\# Trainable\\ Parameters\end{tabular}} & BLEU & NIST & METEOR & ROUGE-L & CIDEr \\ \midrule
\multirow{6}{*}{\rotatebox{90}{\begin{tabular}[c]{@{}c@{}} GPT-2\\ Small\end{tabular}}} & FF        & 123.65M & 65.81 & 8.22 & 45.26 & 71.15 & 2.32 \\
& \multicolumn{1}{l|}{AdapterH}         & \multicolumn{1}{r|}{0.12M}   & 66.11 & 8.35 & 44.39 & 68.75 & 2.39 \\
& \multicolumn{1}{l|}{AdapterL}         & \multicolumn{1}{r|}{0.12M}  & 66.77 & 8.21 & 44.16 & 70.13 & 2.28 \\
& \multicolumn{1}{l|}{FourierFT}         & \multicolumn{1}{r|}{0.017M}  & 66.36 & 8.37 & 45.85 & 70.44 & 2.34 \\
& \multicolumn{1}{l|}{LoRA}      & \multicolumn{1}{r|}{0.13M}   & \textbf{66.94} & 8.32 & 46.26 & 70.97 & 2.33 \\
& \multicolumn{1}{l|}{\textbf{FAA(Ours)}} & \multicolumn{1}{r|}{0.12M}   & 66.56  & \textbf{8.51} & \textbf{46.53} & \textbf{71.51} & \textbf{2.42} \\ \midrule
\multirow{6}{*}{\rotatebox{90}{\begin{tabular}[c]{@{}c@{}} Deepseek\\ R1-
1.5B\end{tabular}}} & FF        & 1.5B & 86.23 & 9.59 & 67.94 & 88.22 & 3.21 \\
& \multicolumn{1}{l|}{AdapterH}         & \multicolumn{1}{r|}{1.63M}   & 86.34 & 9.66 & 68.15 & 88.23 & 2.98 \\
& \multicolumn{1}{l|}{AdapterL}         & \multicolumn{1}{r|}{1.63M}  & 86.75 & 9.67 & 67.76 & 89.48 & 3.13 \\
& \multicolumn{1}{l|}{FourierFT}         & \multicolumn{1}{r|}{0.15M}  & 86.42 & 9.62 & 67.97 & 89.45 & 2.92 \\
& \multicolumn{1}{l|}{LoRA}      & \multicolumn{1}{r|}{1.21M}   & \textbf{87.03} & 9.66 & 68.26 & 88.93 & 3.15 \\
& \multicolumn{1}{l|}{\textbf{FAA(Ours)}} & \multicolumn{1}{r|}{1.64M}   & 76.83  & \textbf{9.69} & \textbf{68.32} & \textbf{89.94} & \textbf{3.22} \\ \midrule
\multirow{6}{*}{\rotatebox{90}{\begin{tabular}[c]{@{}c@{}} LLaMA2\\  7B\end{tabular}}} & FF        & 6.74B & 72.44 & 9.15 & \textbf{50.92} & 74.28 & 2.64 \\
& \multicolumn{1}{l|}{AdapterH}         & \multicolumn{1}{r|}{7.27M}   & 72.72 & 9.26 & 50.33 & 73.94 & 2.62 \\
& \multicolumn{1}{l|}{AdapterL}         & \multicolumn{1}{r|}{7.27M}  & 72.36 & 9.15 & 50.17 & 73.88 & 2.52 \\
& \multicolumn{1}{l|}{FourierFT}         & \multicolumn{1}{r|}{0.82M}  & 72.52 & 9.27 & 49.73 & 73.78 & \textbf{2.74} \\
& \multicolumn{1}{l|}{LoRA}      & \multicolumn{1}{r|}{5.37M}   & 72.41 & 9.32 & 50.27 & 74.38 & 2.67 \\
& \multicolumn{1}{l|}{\textbf{FAA(Ours)}} & \multicolumn{1}{r|}{7.27M}   & \textbf{73.18}  & \textbf{9.33} & 50.23 & \textbf{74.67} & 2.63 \\ \midrule
\multirow{6}{*}{\rotatebox{90}{\begin{tabular}[c]{@{}c@{}} LLaMA3\\  8B\end{tabular}}} & FF        & 8.03B & 82.17 & 9.63 & 61.27 & 83.61 & 3.97 \\
& \multicolumn{1}{l|}{AdapterH}         & \multicolumn{1}{r|}{8.73M}   & 81.79 & 9.47 & \textbf{61.32} & 83.72 & 3.92 \\
& \multicolumn{1}{l|}{AdapterL}         & \multicolumn{1}{r|}{8.73M}  & 82.18 & 9.38 & 61.16 & 83.79 & 3.90 \\
& \multicolumn{1}{l|}{FourierFT}         & \multicolumn{1}{r|}{0.91M}  & 81.98 & 9.57 & 61.27 & 83.65 & 3.99 \\
& \multicolumn{1}{l|}{LoRA}      & \multicolumn{1}{r|}{6.47M}   & \textbf{82.22} & 9.67 & 61.19 & 83.72 & \textbf{4.05} \\
& \multicolumn{1}{l|}{\textbf{FAA(Ours)}} & \multicolumn{1}{r|}{8.73M}   & 82.16  & \textbf{9.72} & 61.16 & \textbf{83.88} & 3.97 \\ \bottomrule
\end{tabular}%
}
\end{table}

\subsubsection{Experimental Results}

Table~\ref{tab:e2e} shows that FAA outperforms other methods on the End-to-End NLG Benchmark. For GPT-2 Small, FAA achieves the highest scores in NIST (8.51), METEOR (46.8), ROUGE-L (71.5), and CIDEr (2.42), with competitive BLEU (66.5). For Deepseek R1-1.5B, FAA leads in NIST (9.69), METEOR (68.32), ROUGE-L (89.94), and CIDEr (3.22), with a BLEU score of 76.8. For LLaMA2-7B, FAA excels in BLEU (73.18), NIST (9.33), and ROUGE-L (74.67). For LLaMA3-8B, our FAA achieves the highest scores in NIST (9.72), and ROUGE-L (83.88), with competitive CIDEr (3.97).

FAA's superior performance is due to its efficient adaptation mechanism using Fourier transforms, which enhances the model's ability to capture complex patterns in text generation. These results demonstrate that incorporating Fourier frequency processing in fine-tuning improves text generation performance, validating the effectiveness of our approach.

\subsection{Instruction Tuning}

\subsubsection{Experimental Setup}

We evaluate instruction tuning by fine-tuning Qwen2-7B, DeepSeek-R1-Distill-Qwen-1.5B, and LLaMA2-7B on five datasets: MT-Bench, Vicuna Eval, BBH, MATH, and Alpaca. MT-Bench, Vicuna Eval, and Alpaca assess conversational ability, while BBH and MATH gauge logical reasoning and mathematical skills. GPT-4 scores MT-Bench and Vicuna Eval (1–10), and LC Win Rate is used for Alpaca. Detailed hyperparameters and training rounds are provided in Appendix ~\ref{APP:Hyperparameter settings}.

\begin{table}
    \small
    \setlength{\tabcolsep}{2pt} 
    \caption{Performance comparison of different methods on the MT-Bench, Vicuna Eval, BBH, MATH, and Alpaca datasets for Qwen2 7B, Deepseek R1-1.5B, LLaMA2-7B and LLaMA3-8B models. We ran 3 experiments with different random seeds and recorded the best test set performance.}
    \label{tab:e2e}
    \resizebox{\columnwidth}{!}{%
    \begin{tabular}{@{}c|l|r|ccccc@{}}
    \toprule
    Model & Method & \multicolumn{1}{c|}{\begin{tabular}[c]{@{}c@{}}\# Trainable\\ Parameters\end{tabular}} & MT-bench & Vicuna Eval & BBH & MATH & Alpaca \\ \midrule
    \multirow{6}{*}{\rotatebox{90}{\begin{tabular}[c]{@{}c@{}} Qwen2\\ 7B\end{tabular}}} & FF        & 7.07B & \textbf{7.88} & 8.88 & 66.74 & 64.11 & 33.72 \\
    & \multicolumn{1}{l|}{AdapterH}         & \multicolumn{1}{r|}{7.29M}   & 7.78 & 8.82 & 66.89 & 64.07 & 33.64 \\
    & \multicolumn{1}{l|}{FourierFT}         & \multicolumn{1}{r|}{0.85M}  & 7.81 & 8.85 & 67.05 & 64.12 & 33.58 \\
    & \multicolumn{1}{l|}{LoRA}      & \multicolumn{1}{r|}{5.40M}   & 7.86 & 8.89 & 67.09 & 64.12 & 33.62 \\
    & \multicolumn{1}{l|}{\textbf{FAA(Ours)}} & \multicolumn{1}{r|}{7.30M}   & 7.82  & \textbf{8.91} & \textbf{67.10}  & \textbf{64.18} & \textbf{33.88} \\ \midrule
    \multirow{6}{*}{\rotatebox{90}{\begin{tabular}[c]{@{}c@{}} Deepseek\\ R1-1.5B\end{tabular}}} & FF        & 1.5B & 8.34 & 8.83 & 88.27 & 84.21 & 71.82 \\
    & \multicolumn{1}{l|}{AdapterH}         & \multicolumn{1}{r|}{1.63M}   & 8.32 & 8.79 & 88.21 & 84.17 & 71.81 \\
    & \multicolumn{1}{l|}{FourierFT}         & \multicolumn{1}{r|}{0.15M}  & 8.33 & 8.82 & 88.07 & 84.23 & 71.86 \\
    & \multicolumn{1}{l|}{LoRA}      & \multicolumn{1}{r|}{1.21M}   & \textbf{8.36} & \textbf{8.85} & 88.17 & 84.16 & \textbf{71.87} \\
    & \multicolumn{1}{l|}{\textbf{FAA(Ours)}} & \multicolumn{1}{r|}{1.63M}   & 8.35  & 8.82 & \textbf{88.29} & \textbf{84.27} & 71.83 \\ \midrule
    \multirow{6}{*}{\rotatebox{90}{\begin{tabular}[c]{@{}c@{}} LLaMA2\\  7B\end{tabular}}} & FF        & 6.94B & 5.19 & 7.39 & 43.67 & 33.21 & 10.87 \\
    & \multicolumn{1}{l|}{AdapterH}         & \multicolumn{1}{r|}{7.27M}   & 5.23 & 7.35 & 43.65 & 33.19 & 10.83 \\
    & \multicolumn{1}{l|}{FourierFT}         & \multicolumn{1}{r|}{0.82M}  & 5.21 & 7.42 & 43.62 & \textbf{33.25} & 10.85 \\
    & \multicolumn{1}{l|}{LoRA}      & \multicolumn{1}{r|}{5.37M}   & 5.22 & \textbf{7.45} & 43.68 & 33.22 & 10.89 \\
    & \multicolumn{1}{l|}{\textbf{FAA(Ours)}} & \multicolumn{1}{r|}{7.27M}   & \textbf{5.24}  & 7.40 & \textbf{43.71} & 33.24 & \textbf{10.92} \\ \midrule
    \multirow{6}{*}{\rotatebox{90}{\begin{tabular}[c]{@{}c@{}} LLaMA3\\  8B\end{tabular}}} & FF        & 8.03B & 8.17 & 8.14 & 56.87 & 46.61 & 30.07 \\
    & \multicolumn{1}{l|}{AdapterH}         & \multicolumn{1}{r|}{8.73M}   & 7.48 & 8.21 & 56.82 & 46.52 & 29.08 \\
    & \multicolumn{1}{l|}{FourierFT}         & \multicolumn{1}{r|}{0.91M}  & 7.41 & 8.23 & 56.85 & 46.57 & 29.79 \\
    & \multicolumn{1}{l|}{LoRA}      & \multicolumn{1}{r|}{6.47M}   & 7.40 & \textbf{8.25} & 56.81 & 46.53 & 30.03 \\
    & \multicolumn{1}{l|}{\textbf{FAA(Ours)}} & \multicolumn{1}{r|}{8.73M}   & \textbf{7.45}  & 8.19 & \textbf{56.89} & \textbf{46.68} & \textbf{30.10} \\ \bottomrule
    \end{tabular}%
    }
\end{table}

\subsubsection{Experimental Results}

The experimental results in Table \ref{tab:e2e} demonstrate the performance of different methods on the MT-Bench, Vicuna Eval, BBH, MATH, and Alpaca datasets for Qwen2 7B, Deepseek R1-1.5B, and LLaMA2 7B models. Our proposed method, FAA (Fourier-Activated Adapter Framework),  outperforms other methods across all datasets and models.
For the Qwen2-7B model, FAA achieves the highest scores in Vicuna Eval (8.91), BBH (67.10), MATH (64.18), and Alpaca (33.88), while maintaining competitive performance in MT-bench (64.18). For the Deepseek R1-1.5B model, FAA leads in BBH (88.29) and MATH (84.27), with strong performance in Vicuna Eval (6.82) and Alpaca (8.83). For the LLaMA2-7B model, FAA excels in MT-bench (5.24), BBH (43.71), and Alpaca (10.92). For the LLaMA3-8B model, FAA achieves the highest scores in MT-bench (7.45), BBH (56.89), MATH (46.68), and Alpaca (30.10).

The superior performance of FAA can be attributed to its efficient and effective adaptation mechanism, which leverages Fourier transforms to enhance the model's ability to capture and process complex patterns in natural language generation tasks. The experimental results demonstrate that the introduction of Fourier frequency processing in fine-tuning can significantly improve the performance of the fine-tuned model, proving the reliability and effectiveness of our model.

\subsection{Frequency perception experiment}

\label{Frequency perception experiment}

\subsubsection{Experimental Setup}
This experiment aims to explore the impact of our FAA on different frequency information in natural language processing tasks. We used five public datasets, including CoLA, WikiText, AG\_News, MRPC, and SST-2, covering tasks such as grammatical understanding, language modeling, news classification, sentence comparison, and sentiment analysis. First, we generated sentence embeddings for each dataset through the pre-trained RoBERTa model and applied Fourier transform to separate the embeddings into high-frequency and low-frequency components. Then, we use FAA to fine-tune these separated datasets to explore the contribution of different frequency components to model performance. 

During fine-tuning, we recorded the L2 norm of 9 Fourier features (num\_grids=9) to assess frequency impact while limiting complexity and plotted heat maps to compare FAA's Fourier weights across different base frequencies. Due to page limitations, we only show the results of CoLA and WikiText in the main text. The results of AG\_News, MRPC, and SST-2 and the specific hyperparameter settings in the experiment are shown in Appendix ~\ref{APP:Hyperparameter settings} and ~\ref{APP:Supplementary experimental results}.
\subsubsection{Experimental Results}
\begin{figure}[t]
    \vspace{-1.5cm}
    \centering
    \includegraphics[width=\columnwidth]{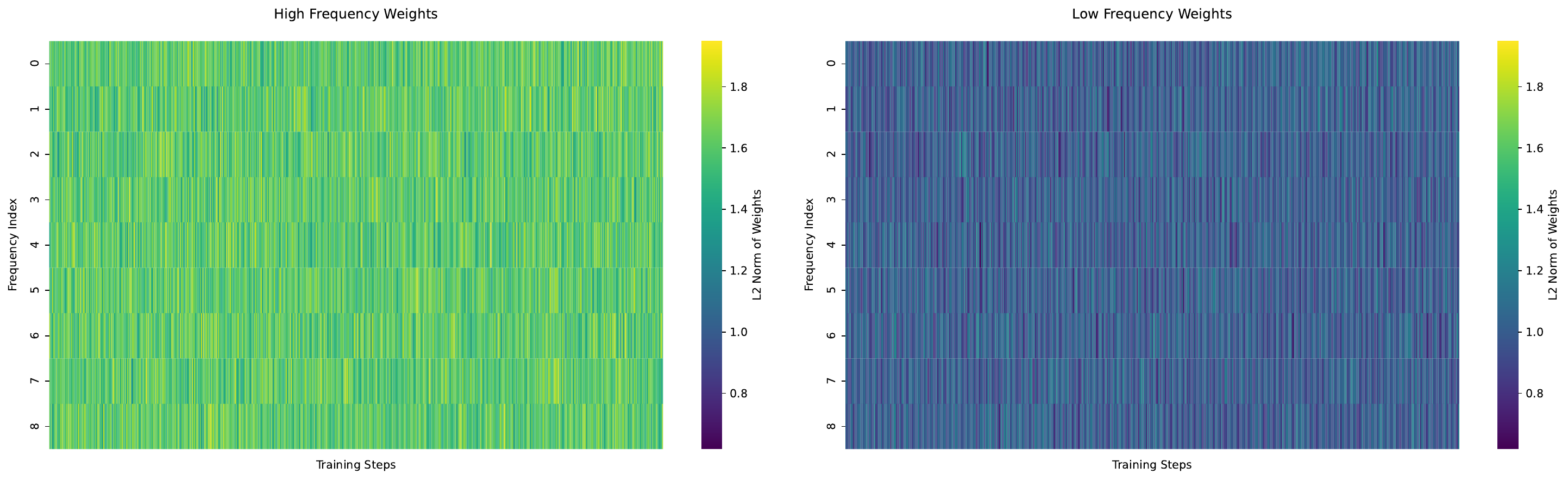}
    \vspace{0.5cm}
    \includegraphics[width=\columnwidth]{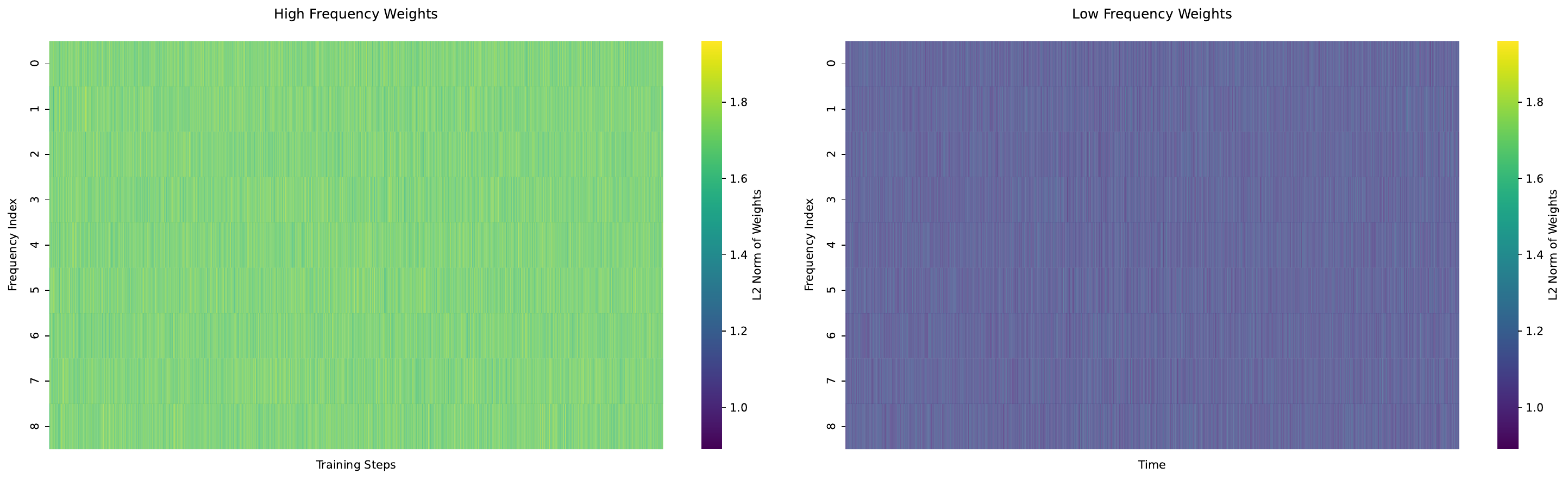}
    \caption{Frequency perception experiment on CoLA (upper) and Wikitext (lower)}
    \label{fig:Frequency}
\end{figure}
Figure \ref{fig:Frequency} shows the $L2$ norm heat maps for CoLA (top) and WikiText (bottom). We observe distinct patterns for high- and low-frequency components, indicating that the Fourier Activation Adapter (FAA) effectively distinguishes different frequency information. High-frequency weights fluctuate more intensely at certain indices, whereas low-frequency weights remain more uniform with lower intensity. This disparity underscores the FAA’s capacity to selectively emphasize or suppress specific frequencies during training.

Moreover, the near-uniform distribution suggests that most frequency components are suppressed, consistent with our L1 regularization \(L_{\text{freq}} = \sum ||r_i||_1\). By enforcing sparsity in the frequency space, this approach reduces complexity and highlights only the most relevant components, ultimately enhancing the model’s performance.

\subsection{Ablation study}

We conducted sufficient ablation experiments to verify the effectiveness of our FAA. Specifically, we conducted fine-tuning experiments from the following five aspects: removing the frequency-aware activation mechanism, removing the adaptive frequency weighting mechanism, unfreezing the RFF internal projection parameters, removing the hierarchical gating mechanism, and hyperparameter selection. Please see Appendix ~\ref{APP:Ablation study} for detailed experimental settings and experimental results.

\section{Conclusion}

In this research, we propose FAA (Fourier-Activated Adapter framework), integrating frequency-domain processing into parameter-efficient fine-tuning. Through introducing random Fourier features and frequency-aware activation mechanisms, FAA enhances the model's ability to capture high-frequency semantic signals. Our evaluations across multiple NLP tasks demonstrate that FAA outperforms traditional adapter methods, with ablation studies validating the importance of adaptive frequency weighting and hierarchical gating. These results highlight the potential of spectral analysis in LLM fine-tuning, advancing research in robust and interpretable adaptation methods.

\section{Limitations}
Despite the promising performance gains, our approach has several limitations. First, compared to mature methods such as LoRA, FAA does not yield a significant reduction in the number of trainable parameters. Second, due to resource constraints, our experiments were conducted on moderately sized datasets and models, and we have not validated the method on larger-scale data or more complex models. Finally, while our work focuses on natural language processing tasks, the application of FAA in other modalities, such as vision and audio, still requires further exploration and empirical validation.
\bibliography{custom}

@misc{brown2020languagemodelsfewshotlearners,
      title={Language Models are Few-Shot Learners}, 
      author={Tom B. Brown and Benjamin Mann and Nick Ryder and Melanie Subbiah and Jared Kaplan and Prafulla Dhariwal and Arvind Neelakantan and Pranav Shyam and Girish Sastry and Amanda Askell and Sandhini Agarwal and Ariel Herbert-Voss and Gretchen Krueger and Tom Henighan and Rewon Child and Aditya Ramesh and Daniel M. Ziegler and Jeffrey Wu and Clemens Winter and Christopher Hesse and Mark Chen and Eric Sigler and Mateusz Litwin and Scott Gray and Benjamin Chess and Jack Clark and Christopher Berner and Sam McCandlish and Alec Radford and Ilya Sutskever and Dario Amodei},
      year={2020},
      eprint={2005.14165},
      archivePrefix={arXiv},
      primaryClass={cs.CL},
      url={https://arxiv.org/abs/2005.14165}, 
}

@misc{zhu2024multilingualmachinetranslationlarge,
      title={Multilingual Machine Translation with Large Language Models: Empirical Results and Analysis}, 
      author={Wenhao Zhu and Hongyi Liu and Qingxiu Dong and Jingjing Xu and Shujian Huang and Lingpeng Kong and Jiajun Chen and Lei Li},
      year={2024},
      eprint={2304.04675},
      archivePrefix={arXiv},
      primaryClass={cs.CL},
      url={https://arxiv.org/abs/2304.04675}, 
}

@misc{bisk2019piqareasoningphysicalcommonsense,
      title={PIQA: Reasoning about Physical Commonsense in Natural Language}, 
      author={Yonatan Bisk and Rowan Zellers and Ronan Le Bras and Jianfeng Gao and Yejin Choi},
      year={2019},
      eprint={1911.11641},
      archivePrefix={arXiv},
      primaryClass={cs.CL},
      url={https://arxiv.org/abs/1911.11641}, 
}

@misc{li2022pretrainedlanguagemodelstext,
      title={Pretrained Language Models for Text Generation: A Survey}, 
      author={Junyi Li and Tianyi Tang and Wayne Xin Zhao and Jian-Yun Nie and Ji-Rong Wen},
      year={2022},
      eprint={2201.05273},
      archivePrefix={arXiv},
      primaryClass={cs.CL},
      url={https://arxiv.org/abs/2201.05273}, 
}

@misc{yuan2023revisitingoutofdistributionrobustnessnlp,
      title={Revisiting Out-of-distribution Robustness in NLP: Benchmark, Analysis, and LLMs Evaluations}, 
      author={Lifan Yuan and Yangyi Chen and Ganqu Cui and Hongcheng Gao and Fangyuan Zou and Xingyi Cheng and Heng Ji and Zhiyuan Liu and Maosong Sun},
      year={2023},
      eprint={2306.04618},
      archivePrefix={arXiv},
      primaryClass={cs.CL},
      url={https://arxiv.org/abs/2306.04618}, 
}

@misc{dodge2020finetuningpretrainedlanguagemodels,
      title={Fine-Tuning Pretrained Language Models: Weight Initializations, Data Orders, and Early Stopping}, 
      author={Jesse Dodge and Gabriel Ilharco and Roy Schwartz and Ali Farhadi and Hannaneh Hajishirzi and Noah Smith},
      year={2020},
      eprint={2002.06305},
      archivePrefix={arXiv},
      primaryClass={cs.CL},
      url={https://arxiv.org/abs/2002.06305}, 
}

@misc{houlsby2019parameterefficienttransferlearningnlp,
      title={Parameter-Efficient Transfer Learning for NLP}, 
      author={Neil Houlsby and Andrei Giurgiu and Stanislaw Jastrzebski and Bruna Morrone and Quentin de Laroussilhe and Andrea Gesmundo and Mona Attariyan and Sylvain Gelly},
      year={2019},
      eprint={1902.00751},
      archivePrefix={arXiv},
      primaryClass={cs.LG},
      url={https://arxiv.org/abs/1902.00751}, 
}

@misc{hendrycks2023gaussianerrorlinearunits,
      title={Gaussian Error Linear Units (GELUs)}, 
      author={Dan Hendrycks and Kevin Gimpel},
      year={2023},
      eprint={1606.08415},
      archivePrefix={arXiv},
      primaryClass={cs.LG},
      url={https://arxiv.org/abs/1606.08415}, 
}

@misc{hu2021loralowrankadaptationlarge,
      title={LoRA: Low-Rank Adaptation of Large Language Models}, 
      author={Edward J. Hu and Yelong Shen and Phillip Wallis and Zeyuan Allen-Zhu and Yuanzhi Li and Shean Wang and Lu Wang and Weizhu Chen},
      year={2021},
      eprint={2106.09685},
      archivePrefix={arXiv},
      primaryClass={cs.CL},
      url={https://arxiv.org/abs/2106.09685}, 
}

@misc{tran2023sparsetopicmodelingspectral,
      title={Sparse topic modeling via spectral decomposition and thresholding}, 
      author={Huy Tran and Yating Liu and Claire Donnat},
      year={2023},
      eprint={2310.06730},
      archivePrefix={arXiv},
      primaryClass={stat.ME},
      url={https://arxiv.org/abs/2310.06730}, 
}

@misc{leethorp2022fnetmixingtokensfourier,
      title={FNet: Mixing Tokens with Fourier Transforms}, 
      author={James Lee-Thorp and Joshua Ainslie and Ilya Eckstein and Santiago Ontanon},
      year={2022},
      eprint={2105.03824},
      archivePrefix={arXiv},
      primaryClass={cs.CL},
      url={https://arxiv.org/abs/2105.03824}, 
}

@misc{verma2024signalprocessinglargelanguage,
      title={Towards Signal Processing In Large Language Models}, 
      author={Prateek Verma and Mert Pilanci},
      year={2024},
      eprint={2406.10254},
      archivePrefix={arXiv},
      primaryClass={cs.CL},
      url={https://arxiv.org/abs/2406.10254}, 
}

@misc{liu2025kankolmogorovarnoldnetworks,
      title={KAN: Kolmogorov-Arnold Networks}, 
      author={Ziming Liu and Yixuan Wang and Sachin Vaidya and Fabian Ruehle and James Halverson and Marin Soljačić and Thomas Y. Hou and Max Tegmark},
      year={2025},
      eprint={2404.19756},
      archivePrefix={arXiv},
      primaryClass={cs.LG},
      url={https://arxiv.org/abs/2404.19756}, 
}

@misc{zhang2025kolmogorovarnoldfouriernetworks,
      title={Kolmogorov-Arnold Fourier Networks}, 
      author={Jusheng Zhang and Yijia Fan and Kaitong Cai and Keze Wang},
      year={2025},
      eprint={2502.06018},
      archivePrefix={arXiv},
      primaryClass={cs.LG},
      url={https://arxiv.org/abs/2502.06018}, 
}

@inproceedings{NIPS2007_013a006f,
  author = {Rahimi, Ali and Recht, Benjamin},
  booktitle = {Advances in Neural Information Processing Systems},
  editor = {J. Platt and D. Koller and Y. Singer and S. Roweis},
  pages = {},
  publisher = {Curran Associates, Inc.},
  title = {Random Features for Large-Scale Kernel Machines},
  url = {https://proceedings.neurips.cc/paper_files/paper/2007/file/013a006f03dbc5392effeb8f18fda755-Paper.pdf},
  volume = {20},
  year = {2007}
}

@misc{xu2023parameterefficientfinetuningmethodspretrained,
      title={Parameter-Efficient Fine-Tuning Methods for Pretrained Language Models: A Critical Review and Assessment}, 
      author={Lingling Xu and Haoran Xie and Si-Zhao Joe Qin and Xiaohui Tao and Fu Lee Wang},
      year={2023},
      eprint={2312.12148},
      archivePrefix={arXiv},
      primaryClass={cs.CL},
      url={https://arxiv.org/abs/2312.12148}, 
}

@misc{liu2019multitaskdeepneuralnetworks,
      title={Multi-Task Deep Neural Networks for Natural Language Understanding}, 
      author={Xiaodong Liu and Pengcheng He and Weizhu Chen and Jianfeng Gao},
      year={2019},
      eprint={1901.11504},
      archivePrefix={arXiv},
      primaryClass={cs.CL},
      url={https://arxiv.org/abs/1901.11504}, 
}

@misc{lv2024parameterfinetuninglargelanguage,
      title={Full Parameter Fine-tuning for Large Language Models with Limited Resources}, 
      author={Kai Lv and Yuqing Yang and Tengxiao Liu and Qinghui Gao and Qipeng Guo and Xipeng Qiu},
      year={2024},
      eprint={2306.09782},
      archivePrefix={arXiv},
      primaryClass={cs.CL},
      url={https://arxiv.org/abs/2306.09782}, 
}

@misc{han2016deepcompressioncompressingdeep,
      title={Deep Compression: Compressing Deep Neural Networks with Pruning, Trained Quantization and Huffman Coding}, 
      author={Song Han and Huizi Mao and William J. Dally},
      year={2016},
      eprint={1510.00149},
      archivePrefix={arXiv},
      primaryClass={cs.CV},
      url={https://arxiv.org/abs/1510.00149}, 
}

@ARTICLE{192463,
  author={Mallat, S.G.},
  journal={IEEE Transactions on Pattern Analysis and Machine Intelligence}, 
  title={A theory for multiresolution signal decomposition: the wavelet representation}, 
  year={1989},
  volume={11},
  number={7},
  pages={674-693},
  doi={10.1109/34.192463}}

@misc{xu2020learningfrequencydomain,
      title={Learning in the Frequency Domain}, 
      author={Kai Xu and Minghai Qin and Fei Sun and Yuhao Wang and Yen-Kuang Chen and Fengbo Ren},
      year={2020},
      eprint={2002.12416},
      archivePrefix={arXiv},
      primaryClass={cs.CV},
      url={https://arxiv.org/abs/2002.12416}, 
}

@inproceedings{He_2023,
   title={Fourier Transformer: Fast Long Range Modeling by Removing Sequence Redundancy with FFT Operator},
   url={http://dx.doi.org/10.18653/v1/2023.findings-acl.570},
   DOI={10.18653/v1/2023.findings-acl.570},
   booktitle={Findings of the Association for Computational Linguistics: ACL 2023},
   publisher={Association for Computational Linguistics},
   author={He, Ziwei and Yang, Meng and Feng, Minwei and Yin, Jingcheng and Wang, Xinbing and Leng, Jingwen and Lin, Zhouhan},
   year={2023},
   pages={8954–8966} }

@misc{hua2025fourierpositionembeddingenhancing,
      title={Fourier Position Embedding: Enhancing Attention's Periodic Extension for Length Generalization}, 
      author={Ermo Hua and Che Jiang and Xingtai Lv and Kaiyan Zhang and Ning Ding and Youbang Sun and Biqing Qi and Yuchen Fan and Xuekai Zhu and Bowen Zhou},
      year={2025},
      eprint={2412.17739},
      archivePrefix={arXiv},
      primaryClass={cs.AI},
      url={https://arxiv.org/abs/2412.17739}, 
}

@misc{jin2024languagemodelssemanticindexers,
      title={Language Models As Semantic Indexers}, 
      author={Bowen Jin and Hansi Zeng and Guoyin Wang and Xiusi Chen and Tianxin Wei and Ruirui Li and Zhengyang Wang and Zheng Li and Yang Li and Hanqing Lu and Suhang Wang and Jiawei Han and Xianfeng Tang},
      year={2024},
      eprint={2310.07815},
      archivePrefix={arXiv},
      primaryClass={cs.IR},
      url={https://arxiv.org/abs/2310.07815}, 
}

@book{+2012,
    url = {https://doi.org/10.1515/9783110274059},
    title = {Volume 1 Frequency Effects in Language Learning and Processing},
    editor = {Stefan Th. Gries and Dagmar Divjak},
    publisher = {De Gruyter Mouton},
    address = {Berlin, Boston},
    doi = {10.1515/9783110274059},
    isbn = {9783110274059},
    year = {2012},
    lastchecked = {2025-02-14}
}

@misc{tamkin2020languageprismspectralapproach,
      title={Language Through a Prism: A Spectral Approach for Multiscale Language Representations}, 
      author={Alex Tamkin and Dan Jurafsky and Noah Goodman},
      year={2020},
      eprint={2011.04823},
      archivePrefix={arXiv},
      primaryClass={cs.CL},
      url={https://arxiv.org/abs/2011.04823}, 
}

@misc{gao2024parameterefficientfinetuningdiscretefourier,
      title={Parameter-Efficient Fine-Tuning with Discrete Fourier Transform}, 
      author={Ziqi Gao and Qichao Wang and Aochuan Chen and Zijing Liu and Bingzhe Wu and Liang Chen and Jia Li},
      year={2024},
      eprint={2405.03003},
      archivePrefix={arXiv},
      primaryClass={cs.LG},
      url={https://arxiv.org/abs/2405.03003}, 
}

@misc{wang2019gluemultitaskbenchmarkanalysis,
      title={GLUE: A Multi-Task Benchmark and Analysis Platform for Natural Language Understanding}, 
      author={Alex Wang and Amanpreet Singh and Julian Michael and Felix Hill and Omer Levy and Samuel R. Bowman},
      year={2019},
      eprint={1804.07461},
      archivePrefix={arXiv},
      primaryClass={cs.CL},
      url={https://arxiv.org/abs/1804.07461}, 
}

@misc{liu2019robertarobustlyoptimizedbert,
      title={RoBERTa: A Robustly Optimized BERT Pretraining Approach}, 
      author={Yinhan Liu and Myle Ott and Naman Goyal and Jingfei Du and Mandar Joshi and Danqi Chen and Omer Levy and Mike Lewis and Luke Zettlemoyer and Veselin Stoyanov},
      year={2019},
      eprint={1907.11692},
      archivePrefix={arXiv},
      primaryClass={cs.CL},
      url={https://arxiv.org/abs/1907.11692}, 
}

@article{Du_ek_2020,
   title={Evaluating the state-of-the-art of End-to-End Natural Language Generation: The E2E NLG challenge},
   volume={59},
   ISSN={0885-2308},
   url={http://dx.doi.org/10.1016/j.csl.2019.06.009},
   DOI={10.1016/j.csl.2019.06.009},
   journal={Computer Speech \& Language},
   publisher={Elsevier BV},
   author={Dušek, Ondřej and Novikova, Jekaterina and Rieser, Verena},
   year={2020},
   month=jan, pages={123–156} 
}

@article{radford2019language,
  title={Language Models are Unsupervised Multitask Learners},
  author={Radford, Alec and Wu, Jeff and Child, Rewon and Luan, David and Amodei, Dario and Sutskever, Ilya},
  year={2019}
}

@misc{deepseekai2025deepseekr1incentivizingreasoningcapability,
      title={DeepSeek-R1: Incentivizing Reasoning Capability in LLMs via Reinforcement Learning}, 
      author={DeepSeek-AI},
      year={2025},
      eprint={2501.12948},
      archivePrefix={arXiv},
      primaryClass={cs.CL},
      url={https://arxiv.org/abs/2501.12948}, 
}

@misc{touvron2023llama2openfoundation,
      title={Llama 2: Open Foundation and Fine-Tuned Chat Models}, 
      author={Meta-AI},
      year={2023},
      eprint={2307.09288},
      archivePrefix={arXiv},
      primaryClass={cs.CL},
      url={https://arxiv.org/abs/2307.09288}, 
}

@misc{grattafiori2024llama3herdmodels,
      title={The Llama 3 Herd of Models}, 
      author={Meta-AI},
      year={2024},
      eprint={2407.21783},
      archivePrefix={arXiv},
      primaryClass={cs.AI},
      url={https://arxiv.org/abs/2407.21783}, 
}

@misc{zheng2023judgingllmasajudgemtbenchchatbot,
      title={Judging LLM-as-a-Judge with MT-Bench and Chatbot Arena}, 
      author={Lianmin Zheng and Wei-Lin Chiang and Ying Sheng and Siyuan Zhuang and Zhanghao Wu and Yonghao Zhuang and Zi Lin and Zhuohan Li and Dacheng Li and Eric P. Xing and Hao Zhang and Joseph E. Gonzalez and Ion Stoica},
      year={2023},
      eprint={2306.05685},
      archivePrefix={arXiv},
      primaryClass={cs.CL},
      url={https://arxiv.org/abs/2306.05685}, 
}

@misc{chiang2023vicuna,
  title        = {Vicuna: An Open-Source Chatbot Impressing GPT-4 with 90\% ChatGPT Quality},
  author       = {Chiang, W.-L. and Li, Z. and Lin, Z. and Sheng, Y. and Wu, Z. and Zhang, H. and Zheng, L. and Zhuang, S. and Zhuang, Y. and Gonzalez, J. E. and others},
  year         = {2023},
  howpublished = {\url{https://vicuna.lmsys.org}},
  note         = {Accessed: 14 April 2023},
}

@article{suzgun2022challenging,
  title={Challenging BIG-Bench Tasks and Whether Chain-of-Thought Can Solve Them},
  author={Suzgun, Mirac and Scales, Nathan and Sch{\"a}rli, Nathanael and Gehrmann, Sebastian and Tay, Yi and Chung, Hyung Won and Chowdhery, Aakanksha and Le, Quoc V and Chi, Ed H and Zhou, Denny and and Wei, Jason},
  journal={arXiv preprint arXiv:2210.09261},
  year={2022}
}

@article{hendrycksmath2021,
  title={Measuring Mathematical Problem Solving With the MATH Dataset},
  author={Dan Hendrycks and Collin Burns and Saurav Kadavath and Akul Arora and Steven Basart and Eric Tang and Dawn Song and Jacob Steinhardt},
  journal={NeurIPS},
  year={2021}
}

@misc{alpaca,
  author = {Rohan Taori and Ishaan Gulrajani and Tianyi Zhang and Yann Dubois and Xuechen Li and Carlos Guestrin and Percy Liang and Tatsunori B. Hashimoto },
  title = {Stanford Alpaca: An Instruction-following LLaMA model},
  year = {2023},
  publisher = {GitHub},
  journal = {GitHub repository},
  howpublished = {\url{https://github.com/tatsu-lab/stanford_alpaca}},
}

@misc{adapterl,
      title={Exploring Versatile Generative Language Model Via Parameter-Efficient Transfer Learning}, 
      author={Zhaojiang Lin and Andrea Madotto and Pascale Fung},
      year={2020},
      eprint={2004.03829},
      archivePrefix={arXiv},
      primaryClass={cs.CL},
      url={https://arxiv.org/abs/2004.03829}, 
}

@misc{adapterp,
      title={AdapterFusion: Non-Destructive Task Composition for Transfer Learning}, 
      author={Jonas Pfeiffer and Aishwarya Kamath and Andreas Rücklé and Kyunghyun Cho and Iryna Gurevych},
      year={2021},
      eprint={2005.00247},
      archivePrefix={arXiv},
      primaryClass={cs.CL},
      url={https://arxiv.org/abs/2005.00247}, 
}

@misc{mahabadi2021compacterefficientlowrankhypercomplex,
      title={Compacter: Efficient Low-Rank Hypercomplex Adapter Layers}, 
      author={Rabeeh Karimi Mahabadi and James Henderson and Sebastian Ruder},
      year={2021},
      eprint={2106.04647},
      archivePrefix={arXiv},
      primaryClass={cs.CL},
      url={https://arxiv.org/abs/2106.04647}, 
}

@misc{huh2024trainingneuralnetworksscratch,
      title={Training Neural Networks from Scratch with Parallel Low-Rank Adapters}, 
      author={Minyoung Huh and Brian Cheung and Jeremy Bernstein and Phillip Isola and Pulkit Agrawal},
      year={2024},
      eprint={2402.16828},
      archivePrefix={arXiv},
      primaryClass={cs.LG},
      url={https://arxiv.org/abs/2402.16828}, 
}

@inproceedings{zheng2024llamafactory,
  title={LlamaFactory: Unified Efficient Fine-Tuning of 100+ Language Models},
  author={Yaowei Zheng and Richong Zhang and Junhao Zhang and Yanhan Ye and Zheyan Luo and Zhangchi Feng and Yongqiang Ma},
  booktitle={Proceedings of the 62nd Annual Meeting of the Association for Computational Linguistics (Volume 3: System Demonstrations)},
  address={Bangkok, Thailand},
  publisher={Association for Computational Linguistics},
  year={2024},
  url={http://arxiv.org/abs/2403.13372}
}

@inproceedings{
z1,
title={{KABB}: Knowledge-Aware Bayesian Bandits for Dynamic Expert Coordination in Multi-Agent Systems},
author={Jusheng Zhang and Zimeng Huang and Yijia Fan and Ningyuan Liu and Mingyan Li and Zhuojie Yang and Jiawei Yao and Jian Wang and Keze Wang},
booktitle={Forty-second International Conference on Machine Learning},
year={2025},
url={https://openreview.net/forum?id=AKvy9a4jho}
}

@inproceedings{
z2,
title={{GAM}-Agent: Game-Theoretic and Uncertainty-Aware Collaboration for Complex Visual Reasoning},
author={Jusheng Zhang and Yijia Fan and Wenjun Lin and Ruiqi Chen and Haoyi Jiang and Wenhao Chai and Jian Wang and Keze Wang},
booktitle={The Thirty-ninth Annual Conference on Neural Information Processing Systems},
year={2025},
url={https://openreview.net/forum?id=EKJhU5ioSo}
}

@inproceedings{Z3,
  title={{CF}-{VLM}: Counterfactual Vision-Language Fine-tuning},
  author={Jusheng Zhang and Kaitong Cai and Yijia Fan and Jian Wang and Keze Wang},
  booktitle={The Thirty-ninth Annual Conference on Neural Information Processing Systems},
  year={2025},
  url={https://openreview.net/forum?id=0qGtaRTsCo}
}

@inproceedings{
z4,
title={{MAT}-Agent: Adaptive Multi-Agent Training Optimization},
author={Jusheng Zhang and Kaitong Cai and Yijia Fan and Ningyuan Liu and Keze Wang},
booktitle={The Thirty-ninth Annual Conference on Neural Information Processing Systems},
year={2025},
url={https://openreview.net/forum?id=YDWRTYgR79}
}

@inproceedings{
Z5,
title={Tri-{MARF}: A Tri-Modal Multi-Agent Responsive Framework for Comprehensive 3D Object Annotation},
author={Jusheng Zhang and Yijia Fan and Zimo Wen and Jian Wang and Keze Wang},
booktitle={The Thirty-ninth Annual Conference on Neural Information Processing Systems},
year={2025},
url={https://openreview.net/forum?id=YGIbwfNWot}
}

@misc{z6,
      title={MM-CoT:A Benchmark for Probing Visual Chain-of-Thought Reasoning in Multimodal Models}, 
      author={Jusheng Zhang and Kaitong Cai and Xiaoyang Guo and Sidi Liu and Qinhan Lv and Ruiqi Chen and Jing Yang and Yijia Fan and Xiaofei Sun and Jian Wang and Ziliang Chen and Liang Lin and Keze Wang},
      year={2025},
      eprint={2512.08228},
      archivePrefix={arXiv},
      primaryClass={cs.CV},
      url={https://arxiv.org/abs/2512.08228}, 
}

@misc{z7,
      title={HybridToken-VLM: Hybrid Token Compression for Vision-Language Models}, 
      author={Jusheng Zhang and Xiaoyang Guo and Kaitong Cai and Qinhan Lv and Yijia Fan and Wenhao Chai and Jian Wang and Keze Wang},
      year={2025},
      eprint={2512.08240},
      archivePrefix={arXiv},
      primaryClass={cs.CV},
      url={https://arxiv.org/abs/2512.08240}, 
}

@misc{z8,
      title={Kolmogorov-Arnold Fourier Networks}, 
      author={Jusheng Zhang and Yijia Fan and Kaitong Cai and Keze Wang},
      year={2025},
      eprint={2502.06018},
      archivePrefix={arXiv},
      primaryClass={cs.LG},
      url={https://arxiv.org/abs/2502.06018}, 
}

@misc{z9,
      title={DrDiff: Dynamic Routing Diffusion with Hierarchical Attention for Breaking the Efficiency-Quality Trade-off}, 
      author={Jusheng Zhang and Yijia Fan and Kaitong Cai and Zimeng Huang and Xiaofei Sun and Jian Wang and Chengpei Tang and Keze Wang},
      year={2025},
      eprint={2509.02785},
      archivePrefix={arXiv},
      primaryClass={cs.CL},
      url={https://arxiv.org/abs/2509.02785}, 
}

@misc{z10,
      title={OSC: Cognitive Orchestration through Dynamic Knowledge Alignment in Multi-Agent LLM Collaboration}, 
      author={Jusheng Zhang and Yijia Fan and Kaitong Cai and Xiaofei Sun and Keze Wang},
      year={2025},
      eprint={2509.04876},
      archivePrefix={arXiv},
      primaryClass={cs.AI},
      url={https://arxiv.org/abs/2509.04876}, 
}

@misc{z11,
      title={Learning Dynamics of VLM Finetuning}, 
      author={Jusheng Zhang and Kaitong Cai and Jing Yang and Keze Wang},
      year={2025},
      eprint={2510.11978},
      archivePrefix={arXiv},
      primaryClass={cs.LG},
      url={https://arxiv.org/abs/2510.11978}, 
}

@misc{z12,
      title={Failure-Driven Workflow Refinement}, 
      author={Jusheng Zhang and Kaitong Cai and Qinglin Zeng and Ningyuan Liu and Stephen Fan and Ziliang Chen and Keze Wang},
      year={2025},
      eprint={2510.10035},
      archivePrefix={arXiv},
      primaryClass={cs.AI},
      url={https://arxiv.org/abs/2510.10035}, 
}

@misc{z13,
      title={Top-Down Semantic Refinement for Image Captioning}, 
      author={Jusheng Zhang and Kaitong Cai and Jing Yang and Jian Wang and Chengpei Tang and Keze Wang},
      year={2025},
      eprint={2510.22391},
      archivePrefix={arXiv},
      primaryClass={cs.CV},
      url={https://arxiv.org/abs/2510.22391}, 
}

@misc{z14,
      title={LLM-CAS: Dynamic Neuron Perturbation for Real-Time Hallucination Correction}, 
      author={Jensen Zhang and Ningyuan Liu and Yijia Fan and Zihao Huang and Qinglin Zeng and Kaitong Cai and Jian Wang and Keze Wang},
      year={2025},
      eprint={2512.18623},
      archivePrefix={arXiv},
      primaryClass={cs.CL},
      url={https://arxiv.org/abs/2512.18623}, 
}

@misc{z15,
      title={DepthSSC: Monocular 3D Semantic Scene Completion via Depth-Spatial Alignment and Voxel Adaptation}, 
      author={Jiawei Yao and Jusheng Zhang and Xiaochao Pan and Tong Wu and Canran Xiao},
      year={2024},
      eprint={2311.17084},
      archivePrefix={arXiv},
      primaryClass={cs.CV},
      url={https://arxiv.org/abs/2311.17084}, 
}

@inproceedings{z16,
    title = "{CCG}: Rare-Label Prediction via Neural {SEM}{--}Driven Causal Game",
    author = "Fan, Yijia  and
      Zhang, Jusheng  and
      Cai, Kaitong  and
      Yang, Jing  and
      Wang, Keze",
    editor = "Christodoulopoulos, Christos  and
      Chakraborty, Tanmoy  and
      Rose, Carolyn  and
      Peng, Violet",
    booktitle = "Findings of the Association for Computational Linguistics: EMNLP 2025",
    month = nov,
    year = "2025",
    address = "Suzhou, China",
    publisher = "Association for Computational Linguistics",
    url = "https://aclanthology.org/2025.findings-emnlp.331/",
    doi = "10.18653/v1/2025.findings-emnlp.331",
    pages = "6243--6256",
    ISBN = "979-8-89176-335-7"
}

@misc{z17,
      title={3DAlign-DAER: Dynamic Attention Policy and Efficient Retrieval Strategy for Fine-grained 3D-Text Alignment at Scale}, 
      author={Yijia Fan and Jusheng Zhang and Kaitong Cai and Jing Yang and Jian Wang and Keze Wang},
      year={2025},
      eprint={2511.13211},
      archivePrefix={arXiv},
      primaryClass={cs.CV},
      url={https://arxiv.org/abs/2511.13211}, 
}

@misc{z18,
      title={Cost-Effective Communication: An Auction-based Method for Language Agent Interaction}, 
      author={Yijia Fan and Jusheng Zhang and Kaitong Cai and Jing Yang and Chengpei Tang and Jian Wang and Keze Wang},
      year={2025},
      eprint={2511.13193},
      archivePrefix={arXiv},
      primaryClass={cs.AI},
      url={https://arxiv.org/abs/2511.13193}, 
}

@misc{z19,
      title={RaCoT: Plug-and-Play Contrastive Example Generation Mechanism for Enhanced LLM Reasoning Reliability}, 
      author={Kaitong Cai and Jusheng Zhang and Yijia Fan and Jing Yang and Keze Wang},
      year={2025},
      eprint={2510.22710},
      archivePrefix={arXiv},
      primaryClass={cs.AI},
      url={https://arxiv.org/abs/2510.22710}, 
}

@article{z20,
author = {Li, Xiaohua and Zhang, Jusheng and Safara, Fatemeh},
title = {Improving the Accuracy of Diabetes Diagnosis Applications through a Hybrid Feature Selection Algorithm},
year = {2021},
issue_date = {Feb 2023},
publisher = {Kluwer Academic Publishers},
address = {USA},
volume = {55},
number = {1},
issn = {1370-4621},
url = {https://doi.org/10.1007/s11063-021-10491-0},
doi = {10.1007/s11063-021-10491-0},
journal = {Neural Process. Lett.},
month = mar,
pages = {153–169},
numpages = {17}
}

@misc{z21,
      title={STORM: Search-Guided Generative World Models for Robotic Manipulation}, 
      author={Wenjun Lin and Jensen Zhang and Kaitong Cai and Keze Wang},
      year={2025},
      eprint={2512.18477},
      archivePrefix={arXiv},
      primaryClass={cs.RO},
      url={https://arxiv.org/abs/2512.18477}, 
}

@misc{z22,
      title={FlashVLM: Text-Guided Visual Token Selection for Large Multimodal Models}, 
      author={Kaitong Cai and Jusheng Zhang and Jing Yang and Yijia Fan and Pengtao Xie and Jian Wang and Keze Wang},
      year={2025},
      eprint={2512.20561},
      archivePrefix={arXiv},
      primaryClass={cs.CV},
      url={https://arxiv.org/abs/2512.20561}, 
}
\appendix

\section{Appendix}

\etocsettocdepth{subsubsection}  
\localtableofcontents

\label{sec:appendix}

\section{Hyperparameter settings}

\label{APP:Hyperparameter settings}
We list the different hyperparameter settings of FAA in the eight tasks of the GLUE benchmark experiment in Table ~\ref{tab:hyperparameters}. The hyperparameters of other fine-tuning methods follow the official settings.

We list the different hyperparameter settings of FAA for different pre-trained large models on the E2E benchmark in Table ~\ref{tab:e2e_hyperparameters_updated}. The best accuracy of the test set in the experiment is recorded. Note that the experiment is based on the fine-tuning platform built by \cite{zheng2024llamafactory}.

We list different hyperparameter settings of FAA for fine-tuning different pre-trained large models on the MT-bench, Vicuna Eval, BBH, MATH, and Alpaca datasets in Table \ref{tab:Cloa paras} and Table \ref{tab:task_hyperparameters}.

We list the hyperparameter settings for fine-tuning RoBERTa Base using our FAA on different high and low-frequency datasets of the GLUE benchmark for frequency-aware experiments in Table \ref{tab:frequency_experiment_hyperparameters}.

\begin{table*}[ht]
    \centering
    \caption{Hyperparameter setup of FAA for the GLUE benchmark.}
    \label{tab:hyperparameters}
    \begin{tabular}{l S S S S S S S S}
    \toprule
    \multirow{2}{*}{Hyperparameter} & \multicolumn{8}{c}{Task} \\
    \cmidrule(lr){2-9}
    & {STS-B} & {RTE} & {MRPC} & {CoLA} & {SST-2} & {QNLI} & {QQP} & {WNLI} \\
    \midrule
    Optimizer & \multicolumn{8}{c}{AdamW} \\
    LR Schedule & \multicolumn{8}{c}{Linear} \\
    Warmup Ratio & \multicolumn{8}{c}{0.06} \\
    num\_grids & \multicolumn{8}{c}{9} \\
    seeds & \multicolumn{8}{c}{\{0, 42,888,1314,1949\}} \\
    Weight Decay &\multicolumn{8}{c}{0.01} \\
    Gradient Clipping & \multicolumn{8}{c}{1.0} \\
    Dropout Rate &\multicolumn{8}{c}{0.1} \\
    \midrule
    Epochs (Base) & 60 & 90 & 30 & 100 & 40 & 40 & 20 & 25 \\
    Learning Rate (FAA) (Base) & 5E-2 & 5E-2 & 5E-2 & 2E-2 & 5E-3 & 5E-2 & 3E-2 & 1E-2 \\
    Learning Rate (Head) (Base) & 9E-3 & 1.1E-2 & 6E-3 & 8E-3 & 6E-3 & 1E-3 & 1E-3 & 1E-3 \\
    Max Seq. Len (Base) & 512 & 512 & 512 & 512 & 512 & 512 & 512 & 512 \\
    Batch Size (Base) & 32 & 32 & 32 & 32 & 32 & 32 & 32 & 32 \\
    Learning Rate Decay (Base) & 0.8 & 0.8 & 0.8 & 0.8 & 0.8 & 0.8 & 0.8 & 0.8 \\
    \midrule
    Epochs (Large) & 30 & 60 & 30 & 80 & 10 & 30 & 20 & 25 \\
    Learning Rate (FAA) (Large) & 7E-2 & 8E-2 & 6E-2 & 4.3E-2 & 4.3E-2 & 6E-2 & 7E-2 & 8E-2 \\
    Learning Rate (Head) (Large) & 1E-3 & 5E-3 & 1E-3 & 1.1E-2 & 1E-3 & 5E-3 & 1E-3 & 5E-3 \\
    Max Seq. Len (Large) & 512 & 512 & 512 & 256 & 128 & 512 & 512 & 512 \\
    Batch Size (Large) & 32 & 32 & 32 & 128 & 32 & 32 & 32 & 32 \\
    Learning Rate Decay (Large) & 0.8 & 0.8 & 0.8 & 0.8 & 0.8 & 0.8 & 0.8 & 0.8 \\
    \bottomrule
    \end{tabular}
\end{table*}

\begin{table*}[ht]
    \centering
    \begin{tabular}{l|cccc}
    \toprule
    \textbf{Hyperparameter} & \textbf{GPT2-Small} & \textbf{DeepSeek-R1-Distill-Qwen-1.5B} & \textbf{LLaMA2-7B} & \textbf{LLaMA3-8B} \\
    \midrule
    Optimizer & \multicolumn{4}{c}{AdamW} \\
    LR Schedule & \multicolumn{4}{c}{Linear} \\
    seeds & \multicolumn{4}{c}{\{0, 10,100,1000,10000,5000,500,50,5,1\}} \\
    \midrule
    Learning Rate (FAA) & 1E-3 & 2E-3 & 3E-3 & 5E-3 \\
    Batch Size & 64 & 128 & 128 & 128 \\
    Weight Decay & 0.01 & 0.02 & 0.02 & 0.03 \\
    num\_grids & 9 & 9 & 9 & 9 \\
    Epochs & 10 & 10 & 10 & 10 \\
    \bottomrule
    \end{tabular}
    \caption{Hyperparameter setup of FAA on the E2E benchmark for different models.}
     \label{tab:e2e_hyperparameters_updated}
\end{table*}

\begin{figure*}[t]
    \centering
    \includegraphics[width=\columnwidth]{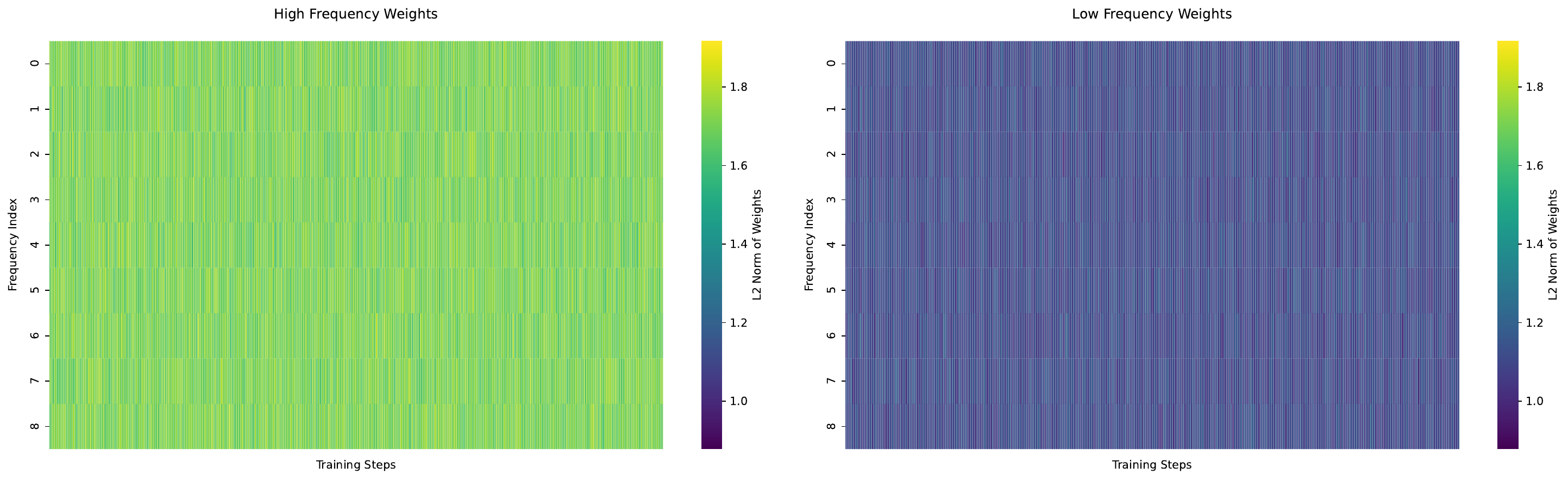}
    \vspace{0.5cm}
    \includegraphics[width=\columnwidth]{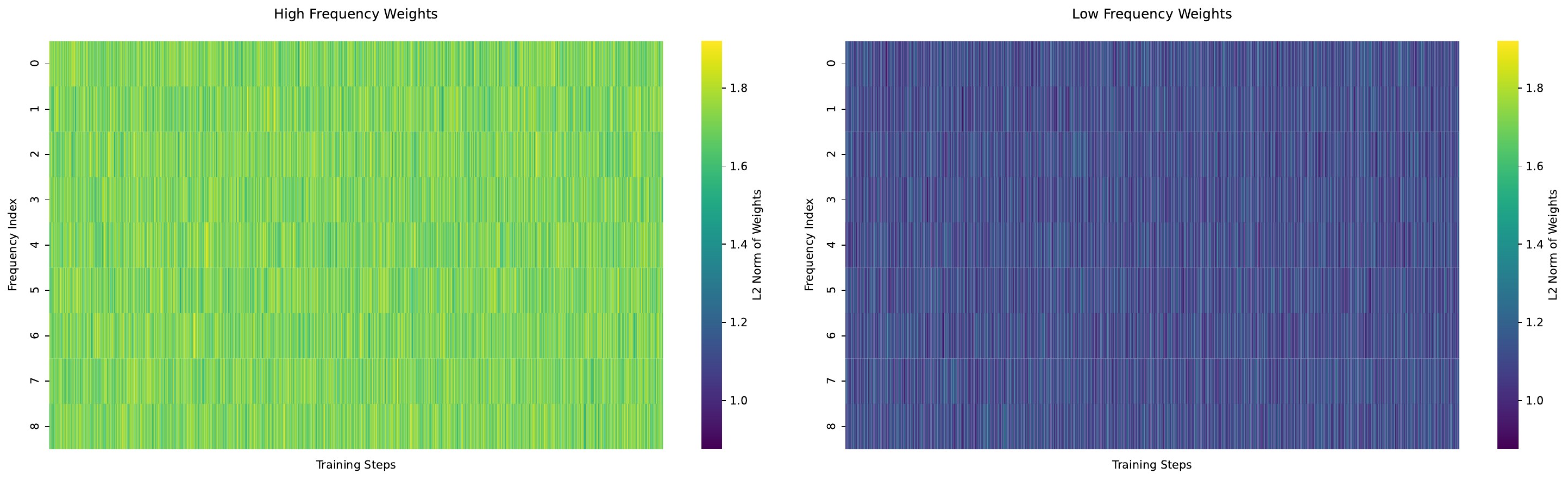}
    \includegraphics[width=\columnwidth]{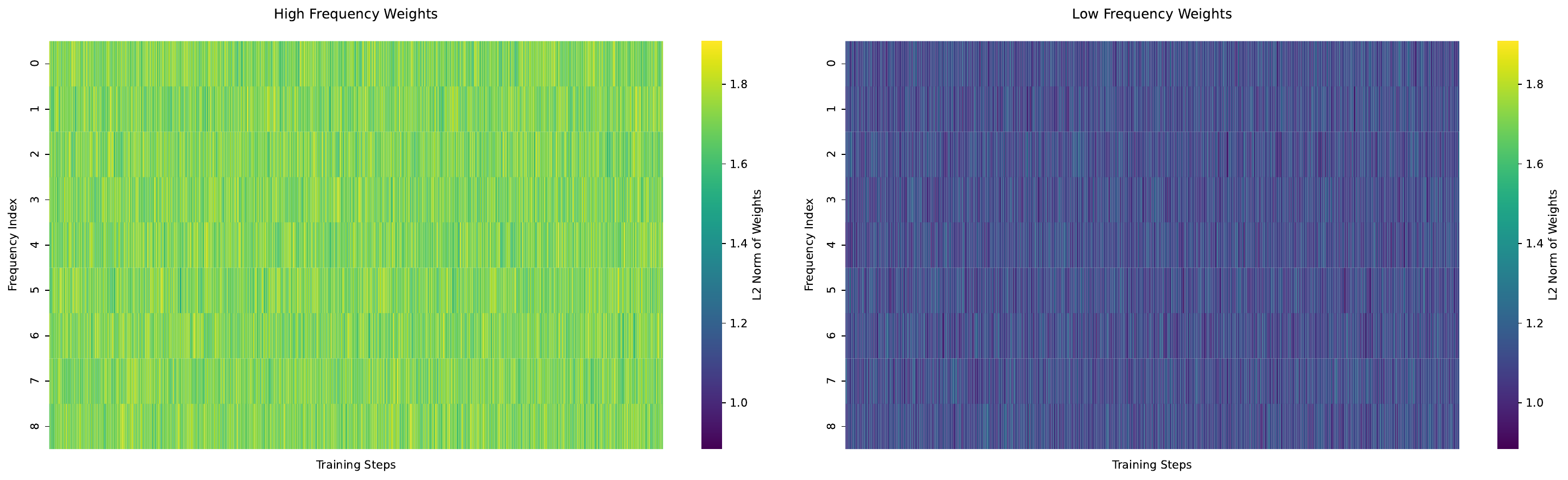}
    \caption{Frequency perception experiment on AG\_NEWS (upper) , MRPC(mid) and SST-2 (lower)}
    \label{fig:Frequency_2}
\end{figure*}

\begin{table*}[ht]
    \centering
    \caption{Hyperparameter setup of FAA on the MT-bench, Vicuna Eval, BBH, MATH, and Alpaca dataset fine-tuning for different models.}
    \label{tab:Cloa paras}
    \begin{tabular}{l|cccc}
    \toprule
    \textbf{Hyperparameter} & \textbf{Qwen2-7B} & \textbf{DeepSeek-R1-Distill-Qwen-1.5B} & \textbf{LLaMA2-7B} & \textbf{LLaMA3-8B} \\
    \midrule
    Optimizer & \multicolumn{4}{c}{AdamW} \\
    LR Schedule & \multicolumn{4}{c}{Linear} \\
    seeds & \multicolumn{4}{c}{\{1000,10000\}} \\
    \midrule
    Weight Decay & 0.01 & 0.02 & 0.02 & 0.03 \\
    num\_grids & 9 & 9 & 9 & 9 \\
    \bottomrule
    \end{tabular}
\end{table*}

\begin{table*}[ht]
    \centering
    \caption{Learning rate and batch size setup of FAA for different models on various tasks. For the number of training rounds, follow the official settings. MT-bench, Vicuna Eval, and BBH are evaluation tools or datasets without a training process, so there are no epoch settings. For the MATH dataset, the epoch is set between 3 and 10, depending on the model and dataset complexity. The official recommendation for Alpaca is to set the epoch to 3.}
    \label{tab:task_hyperparameters}
    \begin{tabular}{l|ccccc}
    \toprule
    \textbf{Task} & \textbf{Qwen2-7B} & \textbf{DeepSeek-R1-Distill-Qwen-1.5B} & \textbf{LLaMA2-7B} & \textbf{LLaMA3-8B} \\
    \midrule
    MT-bench(lr) & 2E-2 & 3E-2 & 4E-2 & 5E-2 \\
    Vicuna Eval(lr) & 1E-3 & 2E-3 & 3E-3 & 4E-3 \\
    BBH(lr) & 5E-2 & 6E-2 & 7E-2 & 8E-2 \\
    MATH(lr) & 1E-2 & 2E-2 & 3E-2 & 4E-2 \\
    Alpaca(lr) & 3E-2 & 4E-2 & 5E-2 & 6E-2 \\
    \midrule
    Batch Size & 32 & 64 & 128 & 256 \\
    \bottomrule
    \end{tabular}
\end{table*}

\begin{table*}[ht]
    \centering
    \caption{Hyperparameter setup for the Frequency perception experiment.}
    \label{tab:frequency_experiment_hyperparameters}
    \begin{tabular}{l|c}
    \toprule
    \textbf{Hyperparameter} & \textbf{Value} \\
    \midrule
    Optimizer & AdamW \\
    LR Schedule & Linear \\
    seeds & \{0, 10, 100, 1000, 10000, 5000, 500, 50, 5, 1\} \\
    Weight Decay & 0.01 \\
    num\_grids & 9 \\
    Epochs & \{CoLA:10,Wikitext:15,AG\_News:5,MRPC:3,SST-2:3\}\\
    Max Seq. Len & 512 \\
    Learning Rate Decay & 0.8 \\
    Attention Heads & 12 \\
    Hidden Layers & 12 \\
    \bottomrule
    \end{tabular}
\end{table*}

\section{Ablation study}

\label{APP:Ablation study}

\subsection{Ablation Experiments}
We conducted a series of ablation experiments to verify the effectiveness of the Fourier Activation Adapter (FAA). Specifically, we explored the following five aspects:

\begin{itemize}
    \item \textbf{Removing the frequency-aware activation mechanism:}This experiment aimed to assess the impact of the frequency-aware activation function on model performance. In this experiment, we remove the parameters that control frequency perception in the model, that is, remove the two learnable parameters \( \alpha^{(l)}, \beta^{(l)} \) in Formula ~\ref{eq:ab}, and examine the performance of the modified FAA fine-tuning model.
    \item \textbf{Removing the adaptive frequency weighting mechanism:}This experiment aimed to evaluate the contribution of the adaptive frequency weighting mechanism. In this experiment, we do not use the adaptive frequency weight adjustment strategy, that is, we do not use Formula ~\ref{eq:Adaptive Frequency Weight Adjustment}. Instead, we set $r_i^{(l)}$ to a learnable parameter initialized using the Xavier strategy and examine the performance of the modified FAA fine-tuning model.
    \item \textbf{Unfreezing the RFF internal projection parameters:}This experiment aims to study the impact of unfreezing the internal projection parameters of random Fourier features (RFF) on the parameters and performance of the model. That is to unfreeze \( W_{\text{rff}} \) in formula \ref{RFF} and make it a trainable parameter.
    \item \textbf{Removing the hierarchical gating mechanism and L1 regularization}: This experiment aimed to determine the impact of the hierarchical gating mechanism and L1 regularization on model performance. Specifically, we remove the loss function in Formula ~\ref{L1} abandon the regularization strategy, and then test the performance of the fine-tuned model.
    \item \textbf{Hyperparameter selection:}This experiment aimed to explore the sensitivity of the model to different hyperparameter settings. We examined the performance of the model with different values of num\_grids to verify why we chose $num\_grids=9$ in most experiments. Note that num\_grids refers to the number of Fourier features, which can also be understood as sampling points in the frequency domain.
\end{itemize}

\subsection{Experimental Setup}

We used five public datasets, including CoLA, QQP, AG\_News, MRPC, and SST-2, covering tasks such as grammatical understanding, paraphrase detection, news classification, sentence comparison, and sentiment analysis. We use the pre-trained RoBERTa base model as the baseline model for fine-tuning using the FAA strategy. The specific experimental hyperparameters are consistent with the NLG experiment.

\begin{table*}[ht]
    \centering
    \caption{Ablation experiment results for different models.}
    \label{tab:ablation_results}
    \begin{tabular}{l|cccccc}
    \toprule
    \textbf{Ablation Experiment} & \textbf{CoLA} & \textbf{QQP} & \textbf{AG\_News} & \textbf{MRPC} & \textbf{SST-2} & \textbf{\#paras} \\
    \midrule
    Original FAA & \textbf{63.3} & \textbf{94.5} & \textbf{95.5} & \textbf{90.2} & \textbf{94.8} & constant\\
    Removing frequency-aware activation & 62.3 & 92.4 & 92.8 & 86.7 & 90.1 & constant\\
    Removing adaptive frequency weighting & 62.9 & 91.2 & 93.6 & 87.8 & 90.7 & constant\\
    Unfreezing RFF internal projection & 62.7 & 91.8 & 94.1 & 88.5 & 91.3 & $+32.1\%$\\
    Removing hierarchical gating and L1 regularization & 56.5 & 89.9 & 92.3 & 85.4 & 89.6 & constant\\
    \bottomrule
    \end{tabular}
\end{table*}

\begin{figure*}[t]  
    \vspace{-1.5cm}
    \centering  
    \includegraphics[width=1\textwidth]{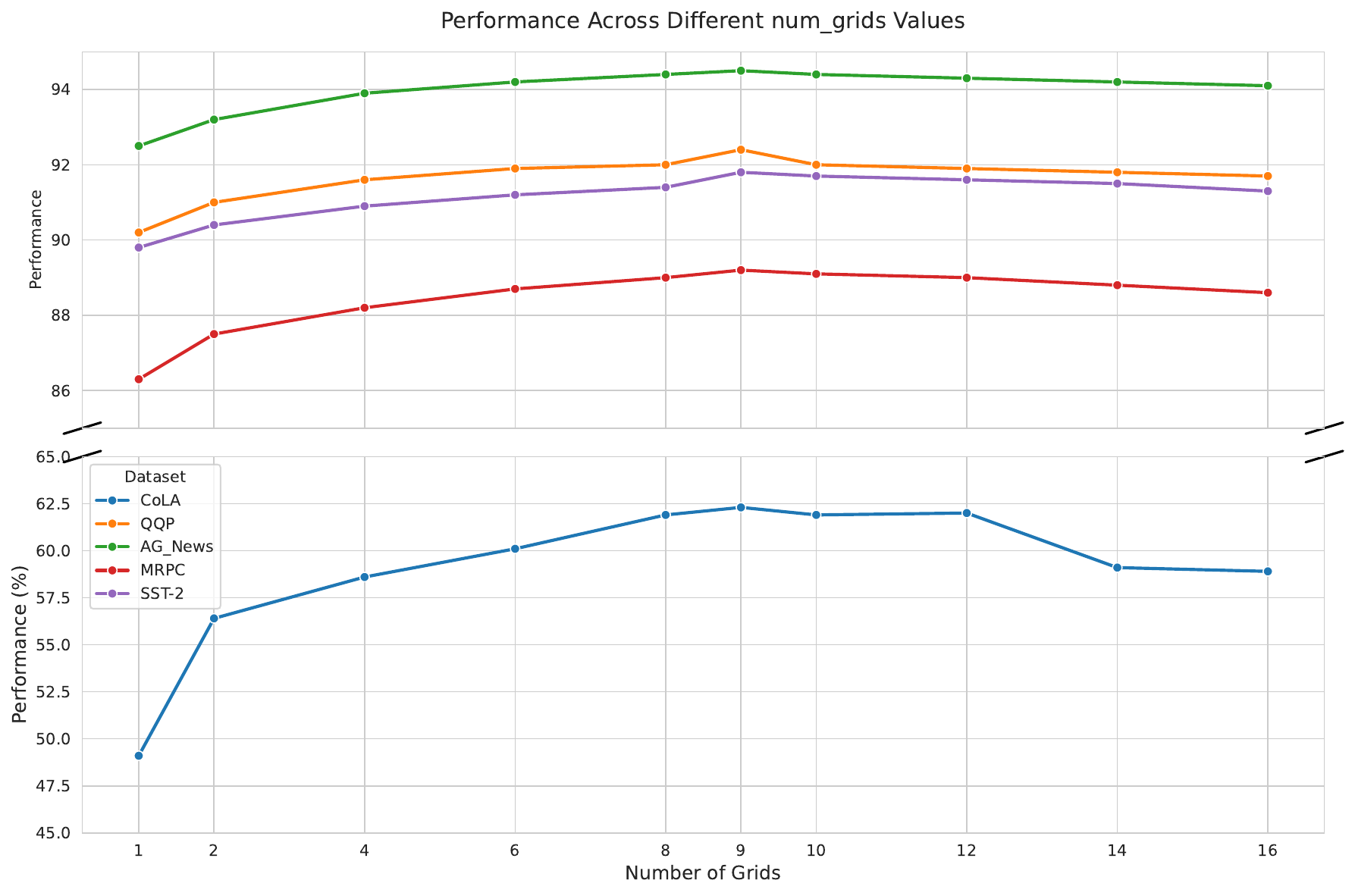}  
    \caption{Performance of different num\_grids values.}
    \label{fig:grids}  
\end{figure*}

\subsection{Experimental Result}

The ablation experiments systematically evaluated the impact of different components of the Fourier Activation Adapter (FAA) on model performance across five diverse NLP tasks: CoLA, QQP, AG\_News, MRPC, and SST-2. The results, as shown in Table \ref{tab:ablation_results}, indicate that removing the frequency-aware activation mechanism led to a noticeable drop in performance across all tasks, with CoLA dropping from 63.3 to 62.3 and MRPC from 90.2 to 86.7. Removing the adaptive frequency weighting mechanism also resulted in performance declines, though less pronounced, with CoLA decreasing to 62.9 and MRPC to 87.8. Unfreezing the RFF internal projection parameters has little effect, with only a slight drop observed on some tasks, but the parameters increase by nearly a third. Removing the hierarchical gating mechanism and L1 regularization led to substantial drops in performance, particularly on CoLA and MRPC, where scores dropped to 56.5 and 85.4, respectively. The hyperparameter selection experiment, as shown in Table \ref{fig:grids}, demonstrated that the number of grids (num\_grids) significantly affects model performance, with num\_grids=9 yielding the best results across most tasks.

Ablation experiments demonstrate the importance and effectiveness of our designed strategy. The frequency-aware activation mechanism is crucial for enhancing the model's ability to capture frequency information, as its removal led to significant performance drops across all tasks. The adaptive frequency weighting mechanism also contributes to performance, though its impact is somewhat mitigated by other components of the model. The internal projection parameters of the RFF do not significantly affect performance, suggesting that their impact is overshadowed by other components. The hierarchical gating mechanism and L1 regularization play critical roles in controlling the flow of information and preventing overfitting, as their removal resulted in substantial performance declines. The hyperparameter selection experiment highlights the importance of choosing an optimal number of grids, with num\_grids=9 providing a good balance between capturing sufficient frequency information and maintaining model generalization. 

\section{Supplementary experimental results}

\label{APP:Supplementary experimental results}

We add some image results of Experiment \ref{fig:Frequency_2} here.

Figure \ref{fig:Frequency_2} illustrates the frequency perception experiment results on AG\_NEWS (upper), MRPC (middle), and SST-2 (lower). The L2 norm heat maps reveal distinct patterns for high- and low-frequency components across these tasks, demonstrating that the Fourier Activation Adapter (FAA) effectively distinguishes different frequency information. In AG\_NEWS, high-frequency weights exhibit more intense fluctuations at specific indices, while low-frequency weights remain relatively uniform with lower intensity. Similarly, in MRPC and SST-2, high-frequency weights show significant variations, whereas low-frequency weights are more stable and less intense. This disparity highlights FAA’s ability to selectively emphasize or suppress specific frequencies during training.

Furthermore, the near-uniform distribution of low-frequency weights suggests that most frequency components are suppressed, aligning with our L1 regularization \(L_{\text{freq}} = \sum ||r_i||_1\). By enforcing sparsity in the frequency space, this approach reduces complexity and highlights only the most relevant components, thereby enhancing the model’s performance. The consistent patterns observed across different tasks underscore the robustness and effectiveness of the FAA in handling various NLP tasks.

\subsection{Datasets and Tasks Overview}
\label{APP:Datasets and Tasks Overview}
In our experiments, we evaluate the performance of FAA fine-tuning across various tasks and datasets. Below is a detailed introduction to each dataset and task used in our study.

\subsubsection{Natural Language Understanding (NLU) Tasks}
We employ the GLUE benchmark, which consists of eight tasks:
\begin{itemize}
    \item \textbf{CoLA}: The Corpus of Linguistic Acceptability is a binary classification dataset that judges the grammaticality of sentences. Each sentence is labeled as either acceptable or not, making it a challenging test for syntactic understanding.
    \item \textbf{SST-2}: The Stanford Sentiment Treebank (SST-2) is used for binary sentiment classification on movie reviews. It provides human-annotated labels that help evaluate a model's capability to capture subjective sentiment nuances.
    \item \textbf{MRPC}: The Microsoft Research Paraphrase Corpus contains pairs of sentences and requires determining whether the two sentences are paraphrases. It challenges models to understand semantic equivalence between different phrasings.
    \item \textbf{QQP}: The Quora Question Pairs dataset consists of pairs of questions and tests whether they are semantically equivalent. This dataset is valuable for assessing a model's ability to detect rephrased or duplicated queries.
    \item \textbf{QNLI}: The Question Natural Language Inference task requires deciding if a sentence contains the answer to a given question. It transforms a question answering task into a binary classification problem, focusing on comprehension.
    \item \textbf{RTE}: Recognizing Textual Entailment (RTE) evaluates whether one sentence logically entails another. This task tests the model's reasoning ability and its understanding of inferential relationships.
    \item \textbf{STS-B}: The Semantic Textual Similarity Benchmark measures the degree of semantic similarity between sentence pairs on a continuous scale. It is used to assess how well models capture subtle semantic nuances.
    \item \textbf{WNLI}: The Winograd Natural Language Inference task is designed around pronoun resolution and requires disambiguating pronouns based on context. It is particularly challenging due to its reliance on subtle linguistic cues.
\end{itemize}

\subsubsection{Natural Language Generation (NLG) Task}
We evaluate the generation capability on the End-to-End NLG benchmark:
\begin{itemize}
    \item \textbf{E2E NLG}: This benchmark is designed for end-to-end natural language generation tasks where models generate textual descriptions from structured inputs. It tests the model's ability to produce coherent, fluent, and accurate text as measured by metrics such as BLEU, NIST, METEOR, ROUGE-L, and CIDEr.
\end{itemize}

\subsubsection{Instruction Tuning Tasks}
For instruction tuning, we fine-tune models on tasks that assess conversational ability, logical reasoning, and instruction following:
\begin{itemize}
    \item \textbf{MT-Bench}: Evaluates the conversational abilities of language models by presenting diverse dialogue scenarios. It measures both the relevance and coherence of generated responses in a conversational setting.
    \item \textbf{Vicuna Eval}: Designed to assess dialogue quality and coherence, it provides a comprehensive evaluation of a model’s ability to maintain context and generate human-like interactions.
    \item \textbf{BBH}: Big-Bench Hard (BBH) focuses on challenging reasoning problems that require complex problem-solving skills, pushing models to demonstrate deeper logical reasoning and inference capabilities.
    \item \textbf{MATH}: The MATH dataset measures the mathematical problem-solving ability of language models through problems that require multi-step reasoning and precise computations.
    \item \textbf{Alpaca}: Evaluates instruction-following performance by testing how well a model adheres to given instructions and generates responses that are contextually appropriate and faithful to the prompts.
\end{itemize}

\subsubsection{Frequency Perception Experiment}
To investigate the impact of frequency information on model performance, we conduct experiments on additional datasets that were not described above:
\begin{itemize}
    \item \textbf{WikiText}: A language modeling dataset containing long-form Wikipedia text. It enables us to study the effects of decomposing sentence embeddings into high- and low-frequency components using the Fourier transform.
    \item \textbf{AG\_News}: A widely-used news classification dataset that categorizes articles into four topics. This dataset allows us to analyze how frequency-aware fine-tuning improves topic discrimination and overall classification performance.
\end{itemize}
\textit{Note: Some data sets have been introduced before and will not be repeated here.}

\subsection{Training Time Analysis}
\label{APP:Training Time Analysis}
To assess the efficiency of our approach, we measured the training time for different fine-tuning methods on the GLUE benchmark using both RoBERTa Base and RoBERTa Large models. We recorded the time per epoch, total training time, and the average number of training steps per second. Table~\ref{tab:training_time_updated} summarizes the results. These measurements help demonstrate that, while our primary focus is on improving performance and frequency perception, our FAA also maintains competitive training efficiency compared to established methods.

\begin{table*}[ht]
\centering
\caption{Training Time Comparison on the GLUE Benchmark.}
\label{tab:training_time_updated}
\begin{tabular}{l l c c c c}
\toprule
\textbf{Method} & \textbf{Model} & \textbf{Epochs} & \textbf{Time per Epoch (min)} & \textbf{Total Time (min)} & \textbf{Steps/sec} \\
\midrule
AdapterH               & RoBERTa Base  & 60              & 6.67                        & 400.0                   & 2.5               \\
LoRA             & RoBERTa Base  & 60              & 5.21                        & 312.6                   & 3.2               \\
FAA (Ours)       & RoBERTa Base  & 60              & 5.05                        & 303.0                   & 3.3               \\
\midrule
AdapterH              & RoBERTa Large & 30              & 7.41                        & 222.3                   & 1.8               \\
LoRA             & RoBERTa Large & 30              & 5.13                        & 153.9                   & 2.6               \\
FAA (Ours)       & RoBERTa Large & 30              & 4.94                        & 148.2                   & 2.7               \\
\bottomrule
\end{tabular}
\end{table*}


\end{document}